\documentclass[conference]{IEEEtran}
\IEEEoverridecommandlockouts
\usepackage{amsmath,amssymb,amsfonts}
\usepackage{algorithmic}
\usepackage{graphicx}
\usepackage{textcomp}
\usepackage{xcolor}
\def\BibTeX{{\rm B\kern-.05em{\sc i\kern-.025em b}\kern-.08em
    T\kern-.1667em\lower.7ex\hbox{E}\kern-.125emX}}

\usepackage{dsbda-style}
\usepackage{booktabs}
\usepackage{threeparttable}

\usepackage{acronym}

\newcommand{\norm}[1]{\left\lVert#1\right\rVert}

\usepackage{booktabs}
\usepackage{amsmath}
\usepackage{amssymb}
\usepackage{placeins}
\usepackage{dsbda-style}
\usepackage{makecell}
\usepackage{comment}
\usepackage{graphicx}

\usepackage[backend=biber,style=ieee,natbib=true]{biblatex} 
 
\AtEveryBibitem{\ifthenelse{\ifentrytype{inproceedings}\OR\ifentrytype{article}\OR\ifentrytype{book}}{ \clearfield{series}\clearfield{volume}\clearfield{number}\clearfield{pages}\clearfield{urldate}\clearfield{doi}}{}}

\usepackage{caption}
\usepackage{subcaption}
\definecolor{lukas}{RGB}{208,240,176}

\begin{document}

\title{Open-World Lifelong Graph Learning}

\author{\IEEEauthorblockN{ Marcel Hoffmann}
\IEEEauthorblockA{
\textit{University of Ulm}\\
Ulm, Germany \\
marcel.hoffmann@uni-ulm.de}
\and
\IEEEauthorblockN{ Lukas Galke}
\IEEEauthorblockA{
\textit{Max Planck Institute for Psycholinguistics}\\
Nijmegen, Netherlands \\
lukas.galke@mpi.nl}
\and
\IEEEauthorblockN{Ansgar Scherp}
\IEEEauthorblockA{
\textit{University of Ulm}\\
Ulm, Germany \\
ansgar.scherp@uni-ulm.de}
}

\maketitle

\begin{abstract}
We study the problem of lifelong graph learning in an open-world scenario, where a model needs to deal with new tasks and potentially unknown classes. We utilize Out-of-Distribution (OOD) detection methods to recognize new classes and adapt existing non-graph OOD detection methods to graph data. Crucially, we suggest performing new class detection by combining OOD detection methods with information aggregated from the graph neighborhood. Most OOD detection methods avoid determining a crisp threshold for deciding whether a vertex is OOD. To tackle this problem, we propose a Weakly-supervised Relevance Feedback (Open-WRF) method, which decreases the sensitivity to thresholds in OOD detection. We evaluate our approach on six benchmark datasets. Our results show that the proposed neighborhood aggregation method for OOD scores outperforms existing methods independent of the underlying graph neural network. Furthermore, we demonstrate that our Open-WRF method is more robust to threshold selection and analyze the influence of graph neighborhood on OOD detection. The aggregation and threshold methods are compatible with arbitrary graph neural networks and OOD detection methods, making our approach versatile and applicable to many real-world applications.
The source code is available at \url{https://github.com/Bobowner/Open-World-LGL}.
\end{abstract}

\section{Introduction}

Lifelong machine learning~\cite{LLL_Def, thrun_lifelong_1998} is a growing research field~\cite{LLL-Images,parisi_continual_2019,Graph_LLL_Survey}, 
where the goal is to update a model to accommodate new data and tasks. 
Most of the research is done on image data~\cite{LLL-Images}, while some work has also been done in the graph domain~\cite{Graph_LLL_Survey}.
A key difference is that images are independently and identically distributed (\iid), while vertices in a graph are edge-connected and thus not independent.
In addition, many graphs change over time, \eg citation graphs, traffic networks, etc., resulting in a distribution shift.
These non-\iid{}\@ characteristics make  updating a graph representation over time a challenging task  in lifelong graph learning (LGL).

Existing work on LGL focuses on the problem of catastrophic forgetting~\cite{LLL_feature_graph,ExpRep_Graph_LLL,Graph_LLL_Survey}.
Catastrophic forgetting is the issue of models forgetting old knowledge, as soon as new data is incorporated~\cite{CatForget}.
Instead, we focus on graph learning that can detect new classes that appear in a graph (\eg new conferences in a citation graph).
While the model incorporates a new class, it ideally maintains its performance on the existing classes.
The task of detecting new classes is a crucial component for open-world learning~\cite{LLL_Def}, 
as opposed to the traditional closed-world setting, where all classes are known during training.

For detecting new classes, the LGL model needs to be able to detect samples that do not belong to the distribution of the training data, \ie are Out-of-Distribution (OOD).
The challenge when classifying new datapoints to OOD versus in-distribution (ID), \ie when performing OOD detection~\cite{OOD_Survey}, is that supervised learning is impossible.
This means that a \textit{threshold} for deciding whether a datapoint is OOD cannot be learned from the data, but must be estimated in the absence of the new classes. 
This challenge is aggravated when data is non-\iid{}\@ like in LGL where the vertices are connected via edges.
So far, existing OOD detection methods on graphs are unsuitable for LGL because they are designed for transductive learning~\cite{uncert_graph_1_GKDE, Uncert_graph_2, OOD_GAT}.
Transductive learning assumes that test nodes are already part of the graph during training.
We address the challenges introduced above by proposing a simple but effective new meta-method for OOD detection, called GOOD (short for: Graph OOD). 
GOOD exploits the homophily assumption of GNNs, \ie adjacent nodes are more likely to belong to the same class, by propagating the OOD scores of existing OOD detection methods through the graph.
To alleviate the thresholding problem of OOD detection, we introduce a weakly-supervised relevance feedback method (Open-WRF) that implements a weakly supervised linear classifier on-top of the raw OOD scores. 
We perform extensive experiments on six benchmark datasets with four GNN models and three OOD detection methods, including ours.
The results show that our meta-method GOOD consistently improves the OOD scores on $4$ out of $6$ datasets, while not reducing them on the others.
{}
Furthermore, we demonstrate that our Open-WRF method for detecting OOD datapoints is robust to the threshold selection.
In summary, our contributions are:

\begin{itemize}
    \item A new meta-method GOOD for OOD detection, which exploits the graph structure to propagate OOD scores (see Section III-d).
    \item A new method Open-WRF that determines robust OOD detection thresholds via weakly-supervised relevance feedback  (see Section III-e). 
    \item Extensive experiments on six benchmark datasets with four GNNs and three OOD detectors demonstrate the effectiveness of our methods. 
\end{itemize}

We discuss related work below.
Our methods are introduced in Section~\ref{sec:Methods}. 
The experimental apparatus is described in Section~\ref{sec:experiment_app}.
The results are presented in Section~\ref{sec:results} followed by a discussion in Section~\ref{sec:discussion}, before we conclude.

\section{Related Work}
\label{sec:rw}
First, we introduce Graph Neural Networks (GNNs) and lifelong learning on GNNs.
Second, we discuss recent OOD detection methods\extended{, which we organize in a-priori methods and post-hoc methods}.

\subsection{GNNs and Lifelong Graph Learning}
\label{sec:rw_gnns+LLL}

\textbf{Graph Neural Networks (GNNs)} 
aggregate features from neighboring vertices and produce a latent representation where the similarity between vertex embeddings corresponds to the similarity of feature vectors and connection patterns in the graph.
Among the most common GNNs are graph convolutional network (GCN)~\cite{GCN}, graph attention network (GAT)~\cite{GAT}, and GraphSAGE~\cite{graph-sage}.
These GNNs rely on explicit message passing.
In contrast, there is Graph-MLP~\citep{Graph-MLP}, which uses the edges of the graph only for a contrastive loss term during training.
\extended{
Following the work of~\citet{Galke_LL}, four families of graph neural networks (GNNs) can be identified: 
isotropic versus anisotropic ~\citep{GNN_categories} and sampling-based versus separating neighborhood aggregation methods. 
Isotropic GNNs treat all edges equally.
Typical representatives are graph convolution networks~\citep{GCN}, GraphSAGE-mean ~\citep{graph-sage}, DiffPool ~\citep{DiffPool} and GIN ~\citep{GIN}.
Anisotropic GNNs apply dynamic edge weights in the aggregation method, e.\,g. GAT ~\citep{GAT}, GatedGCN ~\citep{GatedGCN}, and MoNet ~\citep{MoNet}.
In sampling-based methods, the model applies a certain strategy to sample a subgraph for the model input, mostly based on vertex neighborhood.
Neighborhood separation is based on preprocessing the graph, \eg Simplified Graph Convolution (SGC)~\citep{SimpleGCN} drops the non-linearities of the common GNN architecture, which results in multiplying the adjacency matrix $k$ times to receive a $k$-hop neighborhood.
These GNNs above rely on an explicit message passing, which is aggregating and updating vertex representations across the edges, Graph-MLP~\citep{Graph-MLP} encodes the graph structure in a neighborhood contrastive loss during training.
The neighborhood is defined as a hyperparameter by $r$-th powers of the adjacency matrix.
A beneficial effect is that Graph-MLP does not need access to the graph structure after training.
}
Further details on the GNN models can be found in Appendix~\ref{sec:appendix_models}.

\textbf{Lifelong Machine Learning} (or continual learning) is a continuous learning process~\cite{LLL_Def}, where the learner has to perform a possibly open-ended sequence of tasks $T_1, T_2, T_3$ etc., each with a corresponding dataset $D_1, D_2, D_3 $ etc.
In our setting, the model trains on a task $T_i$ and is evaluated on $T_{i+1}$.
After each task, the knowledge is updated and the lifelong learning process continues with the next task. 
This means that the model is trained on task $T_{i+1}$ using the dataset $D_{i+1}$.
The learner can utilize the knowledge accumulated in previous tasks, which is either implicitly stored in the model parameters trained on the previous tasks or explicitly stored in the datapoints~\cite{Galke_LL}.
A major problem of lifelong learning is catastrophic forgetting~\cite{CatForget}, \ie that old knowledge is quickly lost when the model is trained for new tasks.
A reliable way to alleviate this problem is to keep some datapoints from earlier tasks in the training data~\cite{LLL-Images}.

\extended{
Related to Lifelong Learning are class-incremental learning and data-incremental learning~\citep{LLL_feature_graph}. 
In class incremental learning, the data is split by classes.
This means that for each pair of tasks, $T_i$ and $T_j$, the associated set of possible classes $C_i$ and $C_j$ is disjoint, \ie $C_i \cap C_j = \emptyset$.
The main challenge in class incremental learning is catastrophic forgetting, \ie the same as in lifelong learning when not considering new classes.
Since classes are presented only in one task, the model tends to suffer from catastrophic forgetting after each new task.
In contrast to class-incremental learning, the data-incremental scenario considers tasks where the classes are distributed randomly on the data stream.
The data-incremental setting is equal to the setting of lifelong learning on graphs without the occurrence of new classes.
However, it is not required that all classes are covered by the training/test data in every task.
}

\textbf{Lifelong Graph Learning (LGL)}
is a special case of lifelong machine learning for graphs~\cite{Graph_LLL_Survey}.
\extended{We consider the vertex classification task, where the input is a graph which violates the assumption of independence of most lifelong learning algorithms.}
Methods for LGL have been developed to alleviate the catastrophic forgetting problem by architectural, rehearsal, and regularization approaches~\citep{Graph_LLL_Survey}.
\citet{LLL_feature_graph} follow a different approach by transferring the non-\iid{}\@ vertex classification setting into an independent graph classification setting.
Each vertex of the original graph is transformed to a feature graph, which translates the problem into an (independent) graph classification task.

\extended{
The adjacent vertices of the original graphs are incorporated into the edge weights of the feature graphs by feature vector similarity based on the average inner product of the ego-graph for each vertex.}
\extended{A further challenge and so far rarely studied problem is open-world lifelong graph learning.
The term \emph{open world} indicates the possibility that so far unknown classes may occur in a new task, which need to be detected to enable the model to handle them correctly.}

There are different methods for detecting unknown classes in lifelong learning~\citep{ODIN, IsoMaxPlus, DOC}.
To the best of our knowledge, there is only one in \textbf{open-world} \textbf{lifelong} \textbf{graph} learning~\cite{Galke_LL}.  
The authors extended the open-world classification method Deep Open Classification (DOC)~\citep{DOC} from text classification to graphs (gDOC).
The method tackles imbalanced class distributions with a class-weighted binary cross-entropy loss.
However, it does not use the edges of the graph for the OOD detection, \ie does not exploit all properties of the data.
\extended{A window-based, incremental training procedure is proposed to investigate the trade-off between implicit and explicit knowledge in LGL.}

\subsection{Open-World Learning and Out-of-Distribution Detection}

Following a recent survey on Out-of-Distribution (OOD) detection~\cite{OOD_Survey}, detecting new classes in an unsupervised way, without prior access to the OOD data, is a special case of OOD detection.
As we perform multi-class vertex classification with $N$ classes, where we consider the OOD class as an additional class, $N+1$, our understanding of OOD detection is equivalent to open-set recognition~\cite{OOD_Survey}, a problem where the model is expected to reject samples coming from unknown classes.

Most of the work in OOD detection focuses on the computation of OOD scores~\cite{ODIN, IsoMaxPlus,EnergyScore}. 
Such methods calculate an OOD score $S$ per datapoint based on some statistic of the logits. 
The score $S$ is compared to a threshold $\delta$ to classify a sample as OOD or ID.
Methods that include the determination of a threshold are open-world classification methods~\cite{OOD_Survey}.
In general, OOD detection methods can be distinguished in post-hoc and a-priori methods.
Note that post-hoc is not connected to statistical post-hoc tests in this context.

\textbf{Post-hoc Methods for OOD Detection} are applied to an already trained model, \ie the OOD detector is not part of the main training process.
Post-hoc methods are mostly based on analyses of the output logit distribution.
\extended{
The main benefit of post-hoc methods is that they can be applied to every classification model in a cheap and easy way.
}
A popular example is ODIN~\cite{ODIN}. 
It applies temperature scaling to the logits to adjust the softmax uncertainty of the sample in combination with input perturbations.
ODIN's perturbations have a stronger effect on the ID samples than on the OOD samples and, therefore, help to distinguish the samples better. 
Other popular post-hoc methods are energy-based OOD detection~\cite{EnergyScore} and a common baseline is maximum softmax probability~\cite{OOD_baseline}.
ReAct~\cite{ReAct} is a post-hoc method that requires access to the penultimate network layer and can be combined with other post-hoc OOD detectors\extended{~to further improve the separation of the OOD scores}.
The method is based on the observation that network activations of OOD data are more heterogenous than the ones of ID data.
\extended{
Therefore, ReAct clips the output of the penultimate layer to some hyperparameter $c$. 
The authors suggest to estimate $c$ from the ID data, such that the activation of the ID data is preserved.
This way, the more heterogeneous activation of an OOD sample is clipped in large parts.
This results in a lower mean activation.
The clipped values are passed to the final output layer, where some other OOD method is applied.
}

\textbf{A-priori Methods for OOD Detection} need to be trained together with the model under consideration.
Some of the post-hoc methods described above have been extended to improve the performance of OOD detection for the cost of post-hoc application, such as GenODIN~\citep{GenODIN}.
Further a-priori methods are Deep Open Classifiction (DOC)~\citep{DOC}, IsoMax+ ~\citep{IsoMaxPlus},and DistMax~\citep{DistMax}.
However, none of them is specifically designed for graphs.
It is common for a-priori methods that the output layer of the base model is replaced by a custom layer. 
For example, GenODIN modified ODIN to learn a temperature function for the output layer.
DOC replaced the softmax with a sigmoid layer to determine class-dependent thresholds.
A different approach was proposed with IsoMax+, where the linear output layer is replaced with learnable prototypes in the embedding space.
The OOD scores can be determined by an arbitrary distance measure applied to the prototypes.
\paragraph{Graph-based OOD Detection}
There are only few works on open-world learning and OOD detection explicitly designed for graphs.
\citet{uncert_graph_1_GKDE} used uncertainty estimates based on the vertex path distance distributions encoded in its loss function.
OODGAT~\citep{OOD_GAT} is an approach for OOD detection that relies on the homophily assumption for OOD vertices, \ie OOD vertices and ID vertices form separate clusters in the graph.
For their detection model, they modify the loss function of GAT to OODGAT to learn high edge weights between vertices that both are ID (OOD) and low edge weights between vertices where one is ID and the other ODD.

These OOD detection methods use specialized loss functions to distinguish ID from OOD vertices.
The loss functions are tightly coupled with the GNN model, \ie they are by design not compatible with other GNNs, and expensive to train.
For example, the loss function of OODGAT is only applicable to a GAT graph neural network.
These OOD detection methods assume that the OOD vertices are present during training, \ie the training is inherently transductive. 
Therefore, they are not suitable for LGL, where we consider a growing graph, which inherently requires inductive training.

Finally, OpenWGL~\cite{OpenWGL} is an open-world learning method based on variational auto-encoders (VAE) for static graphs. 
It has been extended by recurrent neural networks to handle snapshots of graphs ~\cite{SeqOpenWGL}.
It is expensive to train due to the use of a VAE.
The graph representation is tightly coupled with the recurrent neural network and the VAE.
Similar to our work, OpenWGL proposes a method to determine thresholds for new class detection, which we use as a baseline in our experiments.

\subsection{Summary}
Overall, we observe that there is little research on open-world learning and OOD detection methods for lifelong graph learning.
To fill this gap, we investigate existing OOD methods and propose to combine OOD scores with information from a vertex neighborhood. 
We empirically evaluate the performance of GNN models and OOD detectors on different graph benchmark datasets, including lifelong learning settings.

\section{OOD Detection Methods and Open-World Learning for Graphs}

\label{sec:Methods}
\extended{
The task of OOD detection can be described as a one-sided one-sample statistical test~\citep{OOD_one_sample_test}, where the hypothesis is that a sample comes from the data distribution $P$ in contrast to some other distribution $Q$.
This means for each sample $x$ the decision is to reject or accept the null hypothesis: 

\begin{align}
    &H_0: x \sim P \\
    &H_A: x \sim Q, P \neq Q
\end{align}
To achieve a crisp decision for the OOD problem, a test statistic $\phi(x)$ is compared to a threshold $\delta$.
Most OOD detection methods compute some OOD score $S(x)$ per sample, which can be seen as the test statistic $\phi$. 
}

We formally introduce the three OOD detection methods used in this paper, namely ODIN, IsoMax+, and gDOC.
Each of them provides an OOD score $S(x)$ for a datapoint $x$.
Subsequently, we present our proposed meta-method GOOD, which performs a weighted combination of OOD scores with the graph structure. The original OOD scores may come from arbitrary OOD detection methods.
Finally, we present our open weakly-supervised relevance feedback (Open-WRF) method to go from OOD detection to open-world learning by estimating OOD detection thresholds.

\paragraph{ODIN}

The OOD detector ODIN~\citep{ODIN} is based on two observations:
First, it is possible to improve the quality of the maximum softmax OOD score~\cite{OOD_baseline} by temperature calibration. 
Second, for an ID sample $x_k$ and an OOD sample $x_l$ with similar OOD scores, \ie $S(x_k) \approx S(x_l)$, the gradient of the ID sample $\norm{ \nabla S(x_k)}$ tends to be larger than for the OOD sample.
Therefore, small perturbations of the input increase the score more for ID samples than for OOD samples.
This results in the following ODIN score, where $S_k$ is computed for each class $k$:

\begin{equation*}
    S_\mathrm{ODIN} = \max_k S_k(\widetilde{x},T) = \max_k \frac{\exp(f_k(\widetilde{x})/T)}{\sum\limits_{l=1}^N \exp(f_l(\widetilde{x})/T)},
\end{equation*}
where $T$ is the temperature hyperparameter, $f_k$ is the model output for class $k$, and $\tilde{x}$ is $x$ perturbed by the perturbation magnitude $\epsilon$ in direction of the negative gradient.
In our setting, $f$ corresponds to a graph-neural network $\operatorname{GNN}(A,X)$, where $A$ is the adjacency matrix and $X$ is the feature matrix of the input graph. 
\extended{
In the original ODIN the regular input $x_k$ is perturbed by:
\begin{equation}
    \widetilde{x} = x - \epsilon \cdot \mathrm{sgn}(-\nabla_x S_y),
\end{equation}
with $S_y = max_k S_k$ and $\epsilon$ as hyperparameters to control the magnitude of the perturbation. 
}

Since GNNs use the adjacency matrix as part of the input, it needs to be perturbed as well.
We do it by changing the non-zero edge weights from $1$ to some other value. 
This results in the following element-wise perturbation of edges in the adjacency matrix $A$ and feature matrix $X$:
\extended{
Since we work on graphs, the perturbation should also consider the neighborhood, \ie the adjacency matrix.
For this reason, we modify the original ODIN and use the whole graph $G=(X,A)$ as input, which is a common approach for GNNs.
Since besides the features $X$ also the adjacency matrix $A$ is part of the input, the strength of the connections will be perturbed as well. 

This can also be seen as a change of edge weights from $1$ to some other value, 
resulting in perturbation equation}

\begin{equation*}
    (\tilde{A},\tilde{X}) = (A,X) - \epsilon \cdot  \mathrm{sgn}(-\nabla_x S_y),
\end{equation*}
with $S_y = \max_k S_k $, $k$ running over the classes and $\nabla_x$ is the gradient with respect to the input sample.
The parameter $\epsilon$ controls the perturbation.
For the adjacency matrix, only non-zero entries are perturbed.

This procedure increases the score $S_{ODIN}(x)$ more for an ID sample than for an OOD sample.
Therefore, the scores for the ID samples tend to be be higher, resulting in a better overall separability. 
\extended{After the procedure, the resulting score $S_{ODIN}(x)$ represents the test statistic for input $x$.}

\extended{The method adjustment for graphs was necessary, because in contrast to images, it is disadvantageous to process each feature vector $x$ of a vertex separately.
This way, neighborhood information would be ignored.}

\paragraph{IsoMax+}
substitutes the output layer by a set of learnable prototypes $p_k$, one for each class.
It uses a cross-entropy loss based on the distance of the vertex embedding to each prototype, where prototypes and embedding have been normalized to length one.
The distance is scaled by a learnable distance scale $d$ and a fixed hyperparameter, the entropic scale $E$, which is removed after training (\ie set to $1$) to adjust for overconfidence \extended{of individual samples}during inference.

\extended{
The common cross-entropy loss is changed to a modified IsoMax+ loss

where $h_x$ is the embedding of the vertex with feature vector $x$, $y_k$ is the vertex label, and $p_k$ the prototype associated with the respective class.

The scaling factors $E_s$ and $d_s$ account for overconfidence in neural networks, and $d_s$ is a learnable distance scale.}

At inference time, one distinguishes class inference and OOD detection.
For class inference, the softmax probability of the prototype distances to each class is used, \ie the closest prototype corresponds to the class.
For OOD detection,~\citet{IsoMaxPlus} showed that the minimum distance score works best:
\begin{equation*}
    S_\mathrm{ISO}(x) =  \min \limits_{k} \norm{\widehat{h}_x-\widehat{p}_k},
\end{equation*}
where $\widehat{h}_x$ and $\widehat{p}_k$ are the length normalized vertex embedding and prototype of vertex $x$ and class $k$.
One drawback of this score is that $S_\mathrm{ISO}(x) \notin [0,1]$.
This requires the threshold to be adjusted to the range of the scores for crisp OOD detection.

\paragraph{gDOC} is an OOD detector for graphs~\citep{Galke_LL}.
It is extended from DOC~\citep{DOC}, which is designed to detect unseen classes in text classification.
gDOC substitutes the final softmax by a sigmoid activation to compute a per-class probability score.
This overcomes the issue of the softmax that it can get arbitrary high confidence scores for meaningless noise inputs~\citep{OverconfidentNN}.
If the logits for all classes fall below a threshold, the sample is considered OOD. 
Therefore, the maximum sigmoid activation can be regarded as the OOD score of DOC and also gDOC, which we use in our experiments. 
Formally, we define
\begin{equation*}
   S_\mathrm{gDOC} = \max_k \left\{ \mathrm{sigmoid}(f_k) \right\}, 
\end{equation*}
where $f_k$ is the logit of the last layer for class $k$.

Unlike the other OOD methods, gDOC also computes an OOD threshold $\delta$.
Since gDOC uses a sigmoid activation, 
each class $k$ has its own threshold $\delta_k$.
To determine these thresholds, gDOC applies a risk reduction method:
The sigmoid output $y \in [0,1]$ for each data point of class $k$ is mirrored by computing $1+(1-y)$ assuming a Gaussian distribution with mean $1$.
The standard deviation $\sigma_k$ of the resulting distribution is estimated and used to determine a threshold by taking the maximum of $\{\delta_{min}, 1- \alpha_\mathrm{DOC} \cdot \sigma_k \}= \delta_k$.

For DOC, the authors choose $\alpha_\mathrm{DOC} = 3$.
\citet{Galke_LL} showed in their extension gDOC that $\alpha_\mathrm{DOC}$ has no effect on imbalanced graph settings.
In gDOC, class weights on the loss function are applied to compensate for imbalanced data in graph learning. 
The weights for each class are computed by $\frac{n-n^+}{n}$, where $n+$ is the number of positive samples of the class and $n$ is the number of all samples in the training dataset.

\paragraph{Proposed Meta-method:~GOOD}
\label{subsubsec:OOD-score-agg}
When naively applying non-graph OOD detection methods to graphs, we miss out on valuable information encoded in the edges of the graph.
GNNs are based on the assumption of graph homophily~\citep{edge-homophily-ratio}, \ie vertices that share an edge tend to share further features and even their class.
Our Graph OOD Detector (GOOD) is a meta-method that combines existing OOD detectors (like those introduced above) with a vertex $v$'s neighborhood information.
GOOD utilizes the graph structure by computing the mean OOD score of $v$'s neighbors.
GOOD then computes the final OOD score with a convex combination:
    
\begin{equation*}
    S_\mathrm{GOOD} = (1-\alpha_\mathrm{OOD}) \cdot S_v + \alpha_\mathrm{OOD} \cdot \frac{1}{|\mathcal{N}(v)|} \sum_{w \in \mathcal{N}(v)} S_w,
\end{equation*}
where $\mathcal{N}(v)$ denotes the neighbors of $v$, $S_v$ is the OOD score before the aggregation, \eg any of the above methods, and $\alpha_\mathrm{OOD}$ is a hyperparameter that controls the influence of the neighbors.
The key benefits of GOOD are that it utilizes the graph structure 
and that it can be combined with any existing OOD method.

\paragraph{Proposed Threshold Determination via WRF}
\label{sec:weakly_supervised_rel_feedback}

Most OOD detectors such as ODIN and IsoMax+ 
focus only on the computation of good OOD scores $S(x)$.
They do not consider the challenge of determining a threshold to decide if a datapoint is OOD, which is necessary for open-world learning.
With Open Weakly-supervised Relevance Feedback (Open-WRF), we propose a method to overcome the challenge of determining a threshold.
We introduce the hyperparameter $q$, which is a domain-interpretable value of the expected ratio of new OOD vertices in the next time step. 
We show that the Open-WRF method is less sensitive to different values of the $q$ parameter than 
the selection of a naive threshold, which can be hard to determine. 

Our Open-WRF method uses scores of an existing OOD detector such as ODIN.
To obtain a crisp detection, we first apply the OOD detector to all inputs in the dataset and compute their OOD scores $S(x_1), \dots, S(x_n)$.
These scores are sorted in ascending order, resulting in a permutation $\pi$, $S(x_{\pi(1)}), \dots, S(x_{\pi(n)})$.
Subsequently, we assign pseudo labels to the datapoints by selecting a domain-specific value for the hyperparameter $q$, \eg $0.05$.
We label the top $q$ percent of the datapoints as OOD and the lower $1-q$ percent as ID, receiving a labeled dataset for OOD detection, \ie $x_{\pi(1)} \dots x_{\pi(n)}$ is labeled with $0_{\pi(1)}, \dots, 0_{\pi(\lfloor n(1-q)\rfloor)},1_{\pi(\lceil n(1-q) \rceil)}, \dots, 1_{\pi(n)}$.
On this labeled dataset, we train an off-the-shelf classifier in a supervised manner to detect OOD samples.
Since we work on graphs, we use a 2-layer GCN~\cite{GCN} as the ID/OOD-classifier for Open-WRF.
The newly introduced hyperparameter $q$ represents the ratio of OOD scores in the domain, which is not only an interpretable number but can also be determined with the help of prior knowledge without extensive tuning.

\paragraph{Summary}
We adapt recent OOD methods to be applicable to graph data.
We normalize all scores such that $S(x) \in [0,1]$, where $0$ is ID and $1$ is OOD.
We propose a meta-method GOOD to incorporate the graph structure and tackle the non-\iid{}\@ nature of the data.
GOOD can be combined with any method that assigns an OOD score to a vertex.
Furthermore, we propose a novel weakly-supervised approach to tackle the threshold problem of OOD detection and make the OOD detectors suitable for open-world learning.
This method alleviates the optimal threshold determination and can be combined with arbitrary OOD detectors. 

\section{Experimental Apparatus}
\label{sec:experiment_app}
We describe the datasets, including their homophily properties.
Subsequently, we explain the experimental procedure, hyperparameter tuning, and performance measures.

\subsection{Datasets: Static and Temporal Graphs}

\paragraph{Description}
\extended{The datasets are divided into two categories: static and temporal, depending on whether they provide time information or not.}
As static datasets, we use the benchmark citation graphs Cora, CiteSeer~\citep{CoraCiteSeer}, and PubMed~\citep{PubMed}.
Descriptive statistics can be found in Table~\ref{tab:datasets}.
We use a random split with 60\% training, 20\% validation, and 20\% test data, sampled with respect to the class distribution.
We also experiment with the Planetoid split~\cite{Planetoid}.
See Section~\ref{sec:discussion} on a reflection of using this split.
To simulate lifelong learning on static graphs, we consider the training data as task $T_0$, validate as $T_1$, and test as $T_2$.

Regarding the temporal datasets, we use three citation graphs dblp-easy and dblp-hard from~\citet{Galke_LL} (both ranging from $1990$ to $2015$) with conferences as labels and OGB-arXiv (from $1971$ to $2020$) from~\citet{ogb-benchmarks} with subject areas as labels.
Each task $T_i$ corresponds to a specific year.
Details can be found in Table~\ref{tab:datasets}.
For dblp-easy, we set $t_0 = 2005$, meaning the first task $T_0$ is trained with data up the year 2005.
For dblp-hard, we set $t_0=2004$.
The values are chosen based on~\citet{Galke_LL} and whether the next year $t_{1}$ has a new class.
The latter is important for hyperparameter tuning.
For OGB-arXiv, $t_0=1997$ since the new occurring classes appear rather early in the timespan.

\begin{table}[!th]
    \centering
    \caption{Global characteristics of  the datasets with the number of vertices $|V|$, edges $\lvert E \rvert$, features $D$, and classes $\lvert \sY \rvert$ along with number of newly appearing classes (in braces) within the $T$ evaluation tasks.}
    \label{tab:datasets}
    
    \small
    
    \begin{tabular}{lrrrrr}
    \toprule
    Static & $\lvert V \rvert$ & $\lvert E \rvert$ & $D$ & $\lvert \sY \rvert$ & $T$\\
    \midrule
    Cora      & $2,708$ & $10,556$ & $1,433$ & $7$\,\,\, $(0)$ & $1$ \\
    CiteSeer      & $3,327$ & $9,104$ & $3,703$ & $6$\,\,\, $(0)$ & $1$ \\
    PubMed     & $19,717$ & $88,648$ & $500$ & $3$\,\,\, $(0)$ & $1$ \\
    \midrule
    Temporal & $\lvert V \rvert$ & $\lvert E \rvert$ & $D$ & $\lvert \sY \rvert$ & $T$\\
    \midrule
    dblp-easy & $45,407$  & $112,131$ & $2,278$ & $12$\,\,\,  $(4)$   & $26$\\
    dblp-hard & $198,675$ & $643,734$ & 4,043 & $73$ $(23)$  & $26$\\
    OGB-arXiv & $169,343$ & $1,166,243$ & $128$ & $40$ $(21)$ & $35$ \\
    \bottomrule
    \end{tabular}

\end{table}

\paragraph{Homophily Measures}
In Section~\ref{sec:Methods}, we proposed a meta-method for OOD scores relying on graph homophily.
We distinguish homophily measures based on the entire graph denoted as graph-level homophily~\citep{edge-homophily-ratio}, on the level of individual vertices~\citep{node-homophily-ratio}, and a class-insensitive version~\citep{class-insensitive-edge-homophily-ratio}.
The equations can be found in Appendix~\ref{sec:homphily_appendix}.
Graph-level homophily is the ratio of edges between two vertices of the same class.
Vertex-level homophily is the average homophily of each vertex with respect to its neighbors.
Class-insensitive homophily is the graph-level homophily computed per class and normalized by the number of classes.
We focus mainly on class-insensitive homophily, since it accounts for the phenomenon that graphs with fewer classes tend to have a higher homophily by chance and produces more comparable measures between datasets with different number of classes.

The homophily scores for the static datasets can be found in Table~\ref{tab:homohily_static}.
More details on homophily per class and information on the connection of homophily to inter- and intra-class edges are shown in Appendix~\ref{sec:homphily_appendix}.
On the temporal datasets, class-insensitive homophily has been computed on the graph for each task and averaged over the time steps.
These homophily scores can be seen in Table~\ref{tab:temp_homophily_avg}.
On the dblp datasets, the time-averaged homophily is higher than the homophily computed on the whole graph. 
We observe that Cora, CiteSeer, PubMed, and OGB-arXiv exhibit a higher degree of homophily than dblp-easy and dblp-hard, which are rather heterophile. 
This and further details and illustrations can be found in Appendix~\ref{sec:ds_appendix}.

\begin{table}[!th]    
    \centering
    \caption{The graph-level, vertex-level, and class-insensitive homophily measures.}
    \label{tab:homohily_static}
    \small
    \begin{tabular}{lrrr}
        \toprule
         Static   & Graph-level & Vertex-level & Class-insens. \\
         \midrule
        Cora        &$0.810$ & $0.825$ & $0.766$ \\
        PubMed      &$0.802$ & $0.792$ & $0.664$ \\
        CiteSeer    &$0.736$ & $0.706$ & $0.627$ \\
        \bottomrule
    \end{tabular}

\vspace{2mm}
    \centering
    \caption{The homophily of the final task and the homophily averaged over each time step along with the standard deviation (in brackets).}
    \label{tab:temp_homophily_avg}
    \small
    \begin{tabular}{lrr}
         \toprule
         Temporal    & \makecell[cc]{ Class-insens. \\ Homophily \\ 
         Last task} & \makecell[cc]{ Class-insens. \\ Homophily \\ Avg over tasks.}  \\

\midrule      
         OGB-arXiv  & $0.444$ & $0.407$ $(\pm0.095)$\\
         dblp-easy  & $0.171$ & $0.254$ $(\pm0.050)$ \\
         dblp-hard  & $0.132$ & $0.190$ $(\pm0.035)$\\ 
         \bottomrule
    \end{tabular}

\end{table}

\subsection{Procedure}

The experimental procedure is divided into two scenarios, one is using the static datasets (Cora, CiteSeer, and PubMed) and the other is using temporal datasets (dblp-easy, dblp-hard, and OGB-arXiv).
We use the GNNs GCN, GraphSAGE, GAT, and Graph-MLP. 
As OOD detectors, we use ODIN, IsoMax+, gDOC, and our GOOD method introduced in Section~\ref{sec:Methods}.
We combine each GNN with each OOD detection method.
To evaluate the threshold methods, we select the GNN model and OOD detector combination that has the best AUROC scores for each dataset.
We compare our Open-WRF method to gDOC and the threshold determination of OpenWGL~\cite{OpenWGL}.
OpenWGL determines the threshold by calculating the average maximum probability of the logits from all samples and the average maximum probability of the logits from $10\%$ of the samples with the largest entropy. 
Then they sum up both numbers and divide them by $2$.
We also include a naive application of thresholds, \ie comparing the scores $S(x)$ to a (fixed) value of $\delta$, which is typically done with OOD detection methods.

For the \textbf{static datasets}, we simulate a three-step lifelong learning procedure.
We use the dataset's training graph as task $T_0$ (without unseen classes), $T_{1}$ for validation, and $T_{2}$ as the test graph including the new classes.
To simulate an OOD setting, we leave out one class $k$ and treat it as the OOD class.
To this end, we remove all the vertices of class $k$ and their edges in $T_0$ and $T_1$. 
On task $T_2$, we evaluate whether each vertex is correctly classified or correctly detected as OOD.
We use a random $60\%$ train, $20\%$ validate, and $20\%$ test split.
We remove test and OOD vertices from the train graph.
Thus, we train our models inductively, which best reflects a lifelong learning setting.
We repeat this procedure for all classes (cross-validation) and average the results.

For the \textbf{temporal datasets}, the graphs are divided into subgraphs according to the time stamp of the vertices.
In our case, the time steps are years. 
Therefore, we end up with test tasks $T_1, \dots, T_N$ corresponding to time steps $t_1, \dots, t_N $.
For each dataset, the model is first trained up to time step $t_0$ and evaluated on $t_1$ to tune the hyperparameters of the GNN models and the OOD detector. 
The GNNs and OOD detectors are then trained for each task up to time step $t_i$ and evaluated on the graph for time step $t_{i+1}$.
Past vertices are always used in every task, \ie we use the whole graph history for training~\citep{Galke_LL}, to avoid effects of catastrophic forgetting.
This means that the last task $T_N$ uses the entire graph except the last year for training.

\subsection{Hyperparameter Optimization}

The hyperparameters of the GNN models are tuned on the validation set without OOD classes.
This is necessary to obtain a good classifier, since the OOD detection performance is highly influenced by the classifier performance~\cite{GoodClassifier}. 
We tune the OOD detection hyperparameters after the GNN models are optimized.
However, for the OOD detectors IsoMax+ and gDOC, which use different output layers, we conduct a combined optimization, in contrast to ODIN, which can be applied to already trained GNNs.
For details, we refer to Appendix~\ref{sec:hp_appendix}.

The hyperparameters of the OOD methods are tuned for each GNN model on each dataset separately.
For ODIN, we tune the temperature $T$ and perturbation magnitude $\eps$. 
gDOC has no hyperparameter that is relevant for comparing OOD scores.
IsoMax+ has the entropic scale $E$ parameter, which is set to $10$ during training as proposed by~\citet{IsoMaxPlus}.
DOC has an extra $\alpha_\mathrm{DOC}$ parameter for threshold determination, where we set $\alpha_\mathrm{DOC} =3$ following~\citet{Galke_LL}.
For our meta-method GOOD, we choose the OOD detector with the best AUROC scores on the validation set, \ie on time step $t_1$.
On this time step, we also determine the  $\alpha_\mathrm{OOD}$ parameter. 
We test $\alpha_\mathrm{OOD} \in \{0.0, 0.1 \dots, 1.0\}$.
The GNN model and OOD method combination that achieves the best AUROC score is used to evaluate the threshold methods.
We use a fixed threshold $\delta =0.1$ for OpenWGL as proposed by~\citet{OpenWGL}.
For a fair comparison, we use $\delta_{min} = 0.1$ for gDOC and $0.1$ for the naive threshold as well.
The same holds for Open-WRF, where we set $q=0.1$, \ie assume that there are 10\% of new vertices that are OOD and belong to a new class.
For the final hyperparameter values of each model we refer to Appendix~\ref{sec:hp_appendix}.

\subsection{Measures}

In both static and temporal settings, we compute the accuracy for vertex classification on the known classes.
Therefore, the test accuracy describes the in-distribution accuracy.
For OOD detection, we need to distinguish between threshold-free and threshold-dependent metrics.
In the threshold-free OOD detection, we report the area under receiver operator characteristic (AUROC), which gives a measure of the produced scores of an OOD detector.
We measure the threshold-dependent OOD detection performance by the micro F1 score on the ID vs. OOD classification.
The scores are averaged over classes for the static datasets and over time for the temporal datasets.

\section{Results: OOD Detection and Classification}
\label{sec:results}

\paragraph{OOD Detection} 
The results for each GNN model and OOD detector combination applied on each dataset can be found in Table~\ref{tab:main_results}.
The left side of the '/'-symbol shows the test accuracy and the right side shows the AUROC score.
Note that the test accuracy of ODIN is always equal to an accuracy score of a model without any OOD detection.
This is because ODIN is a post-hoc method and does not influence the classification behavior of the model, neither during training nor inference.  
On the homophile datasets, our method GOOD is able to improve OOD scores in all settings by at least a small margin and never reduces them. 

We observe that the OOD detector gDOC performs best in all combinations on the temporal datasets, except for dblp-hard with GCN and OGB-arXiv with GAT, where it is outperformed by ODIN.
On the static datasets, IsoMax+ outperforms all OOD detectors when combined with Graph-MLP.
It shows low performance on the temporal graphs OGB-arXiv and dblp-easy (experiments with dblp-hard were omitted, since it is a more challenging version of dblp-easy). 
Overall, we conclude that IsoMax+ and Graph-MLP is the best combination for static datasets, but there is no clear best OOD method and GNN combination for temporal datasets.

\setlength{\tabcolsep}{4pt}
\begin{table}[!th]
\caption{The test accuarcy/AUROC results for each OOD detector, model, and dataset combination. The best AUROC score for each GNN and dataset is marked in bold. The value of the OOD method used for GOOD is underlined.}
\label{tab:main_results}

\begin{tabular}{lrrrr}
     \toprule
       &GCN      &GraphSAGE     &GAT        &Graph-MLP  \\
      \midrule
      \textbf{Static} & Acc/AUROC &  Acc/AUROC &  Acc/AUROC &  Acc/AUROC \\
      \midrule
      Cora&&&&\\
     \midrule
     ODIN&  $0.89$/$0.84$ & $0.89$/$0.82$  & $0.88$/\underline{$0.86$}  & $0.90$/$0.91$\\
    IsoMax+& $0.88$/$0.78$  & $0.85$/$0.75$  & $0.89$/$0.81$  &$0.89$/\underline{$0.95$}\\
     gDOC&  $0.88$/\underline{$0.84$} & $0.88$/\underline{$0.85$}  & $0.88$/$0.84$  &$.0.90$/$0.95$\\
     GOOD (own) & $0.89$/$\mathbf{0.86}$  & $0.89$/$\mathbf{0.89}$  & $0.89$/$\mathbf{0.88}$  &$0.91$/$\mathbf{0.98}$\\
     \midrule
      CiteSeer&&&& \\
     \midrule
     ODIN& $0.77$/\underline{$0.77$} & $0.76$/$0.77$   & $0.78$/\underline{$0.80$} & $0.90$/$0.93$\\
     IsoMax+&  $0.78$/$0.70$ & $0.76$/$0.70$  & $0.78$/$0.76$  & $0.87$/\underline{$\mathbf{0.96}$}\\
     gDOC& $0.77$/$0.76$ & $0.78$/\underline{$0.80$}  & $0.77$/$0.76$  & $0.87$/$0.93$\\
    GOOD (own) & $0.76$/$\mathbf{0.79}$  & $0.78$/$\mathbf{0.81}$  & $0.79$/$\mathbf{0.82}$  &$0.879$/$\mathbf{0.96}$\\
     \midrule
     PubMed&&&& \\
     \midrule
     ODIN& $0.92$/\underline{$0.57$}  & $0.93$/$0.53$  &$0.92$/$0.54$   & $0.92$/$0.62$\\
     IsoMax+& $0.93$/$0.56$  &$0.92$/$0.55$  & $0.92$/\underline{$0.55$}  & $0.91$/\underline{$\mathbf{0.96}$}\\
     gDOC& $0.92$/$0.53$ & $0.93$/\underline{$0.56$} &$0.92$/$0.53$  & $0.89$/$0.95$\\
     GOOD (own) & $0.92$/$\mathbf{0.62}$  & $0.93$/$\mathbf{0.57}$  & $0.92$/$\mathbf{0.57}$  & $0.90$/$\mathbf{0.96}$ \\
     \midrule
     \textbf{Temporal}  & Acc/AUROC &  Acc/AUROC &  Acc/AUROC &  Acc/AUROC \\
     \midrule
     OGB-arXiv&&&& \\
     \midrule
     ODIN& $0.29$/$0.50$  & $0.50$/$0.59$   & $0.45$/\underline{$0.57$}   & $0.66$/$0.52$\\
     IsoMax+ & $0.04$/$0.49$   &$0.10$/$0.50$   & $0.04$/$0.48$   & $0.103$/$0.50$\\
     gDOC& $0.55$/\underline{$0.54$} & $0.51$/$\underline{\mathbf{0.62}}$  & $0.43$/$0.54$  &$ 0.52$/\underline{$0.60$}\\
     GOOD (own) & $0.55$/$\mathbf{0.58}$  & $0.51$/$0.61$  &  $0.44$/$\mathbf{0.61}$  &$0.66$/$\mathbf{0.67}$\\
     \midrule
     dblp-easy&&&& \\
     \midrule
     ODIN & $0.32$/$0.53$  & $0.62$/$0.60$   & $0.43$/$0.58$   & $0.65$/$\mathbf{0.68}$\\
     IsoMax+ & $0.07$/$0.50$   &$0.04$/$0.51$   &$0.10$/$0.49$   & $0.08$/$0.49$\\
     gDOC& $0.37$/\underline{$\mathbf{0.57}$} &$ 0.62$/$\underline{0.63}$  & $0.43$/\underline{$\mathbf{0.60}$}  & $0.66$/\underline{$\mathbf{0.68}$}\\
     GOOD (own) & $0.37$/$0.56$  & $0.62$/$\mathbf{0.64}$  & $0.43$/$\mathbf{0.60}$   & $0.663$/$\mathbf{0.68}$\\
     \midrule
     dblp-hard&&&& \\
     \midrule
     ODIN&  $0.31$/\underline{$\mathbf{0.59}$} & $0.35$/$0.51$   & $0.33$/$\mathbf{0.59}$   &$0.58$/$0.54$\\
     gDOC& $0.30$/$0.53$ & $0.35$/\underline{$\mathbf{0.55}$}  & $0.31$/\underline{$0.56$}  & $0.42$/\underline{$0.55$}\\
     GOOD (own) & $0.31$/$0.54$  & $0.35$/$\mathbf{0.55}$   & $0.31/0.56$  & $0.42$/$\mathbf{0.56}$\\
     
     \bottomrule
\end{tabular}
\setlength{\tabcolsep}{6pt}

\end{table}

\begin{table}[!ht]
    \centering
    \caption{The micro F1 scores of the new class detection in the open world classification setting after the threshold has been applied. The best value for each dataset is marked in bold. Note that we use only the threshold method of OpenWGL, since all methods are applied to the best performing model regarding the AUROC scores.}
    \label{tab:ood_crisp}
    \begin{tabular}{lrrrr}
        \toprule
         &  Open-WRF (own) & gDOC & OpenWGL & Naive \\
         \midrule
         \textbf{Static} &&&\\
         \midrule
        Cora & $\mathbf{0.717}$ & $0.466$ &$0.143$  &$0.143$ \\
        CiteSeer & $\mathbf{0.731}$ & $0.460$ &$0.167$  &$0.167$\\
        PubMed & $\mathbf{0.451}$ & $0.413$ &$0.333$  &$0.333$ \\
        \midrule
        \textbf{Temporal} &&& \\
        \midrule
        OGB-arXiv & $\mathbf{0.982}$ & $\mathbf{0.982}$  &$\mathbf{0.982}$ &$\mathbf{0.982}$ \\
        dblp-easy &$\mathbf{0.984}$& $\mathbf{0.984}$ &$\mathbf{0.984}$ &$\mathbf{0.984}$\\
        dblp-hard & $\mathbf{0.495}$ & $0.491$ &$0.017$ &$0.017$\\
        \bottomrule
    \end{tabular}

\end{table}

\paragraph{Open World Classification}
The results for the thresholding methods can be found in Table~\ref{tab:ood_crisp}.
We observe that Open-WRF is better or equal to the baseline methods for a fixed threshold $\delta=0.1$ (or $q=0.1$) across all datasets.
We see that OpenWGL and naive thresholding always achieve the same scores since they classify all samples in the same class as either OOD or ID based on the dataset being used.
Both are outperformed by gDOC and Open-WRF, which do not suffer from this problem.
However, the same issue occurs for Open-WRF and gDOC on the temporal datasets, except for dblp-hard where they outperform the other two baselines.

\begin{figure*}[!ht]
     \centering
     \begin{subfigure}[b]{0.32\textwidth}
         \centering
        \includegraphics[width = \textwidth]{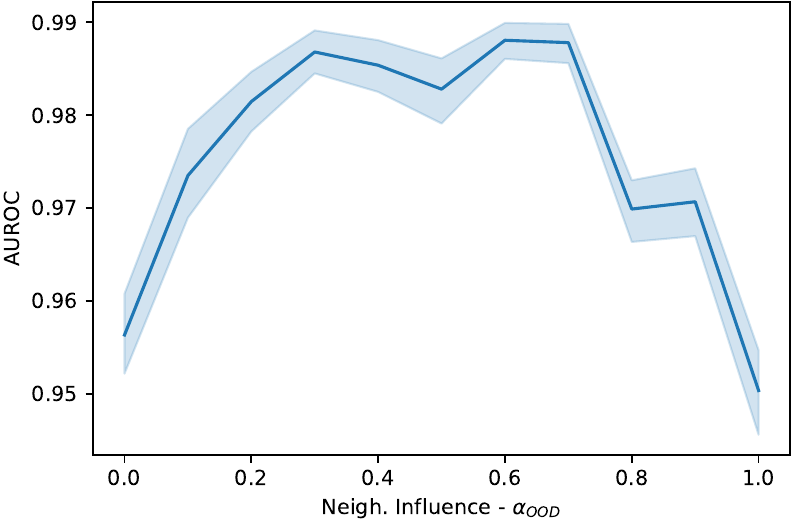}
        \caption{Cora}
        \label{fig:cora_neigh_inf_622}
     \end{subfigure}
     \hfill
     \begin{subfigure}[b]{0.32\textwidth}
         \centering
         \includegraphics[width =\textwidth]{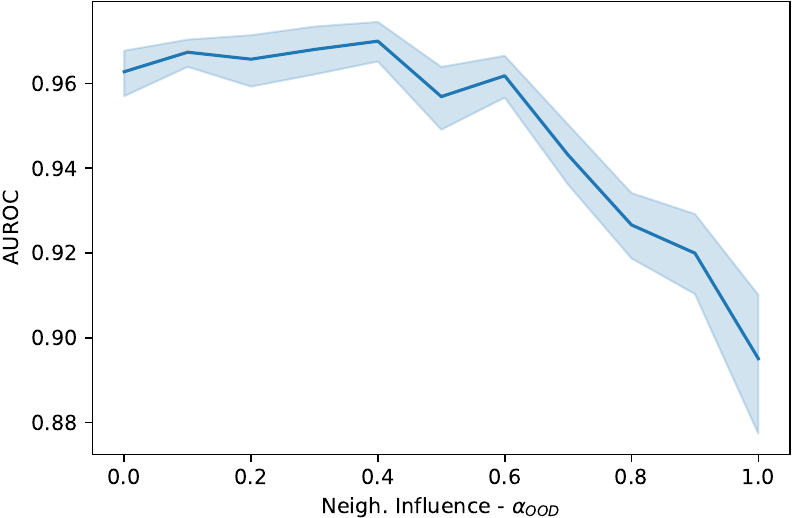}
        \caption{CiteSeer}
        \label{fig:citeseer_neigh_inf_622}
     \end{subfigure}
     \hfill
     \begin{subfigure}[b]{0.322\textwidth}
         \centering
         \includegraphics[width =\textwidth]{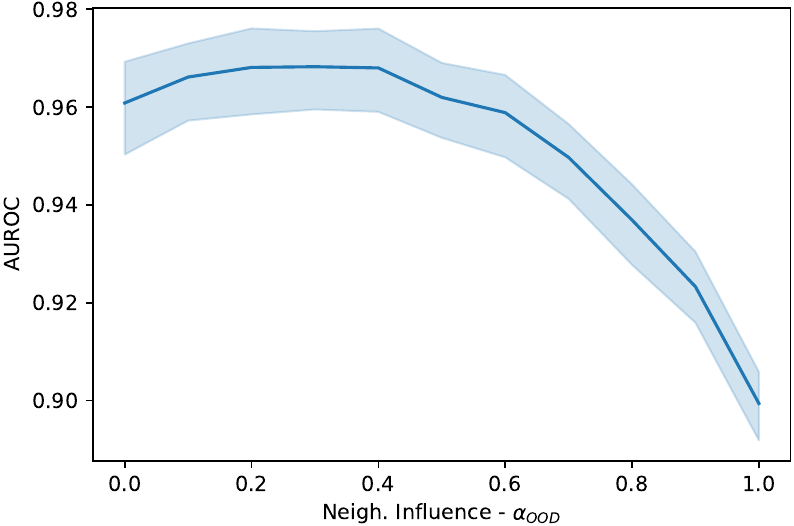}
    \caption{PubMed}
    \label{fig:pubmed_neigh_inf_622}
     \end{subfigure}
     \caption{Measurement of the influence of the $\alpha_\mathrm{OOD}$ values and the respective AUROC scores for the static datasets.}
    \label{fig:neighinfluence_static_datasets_622}
\end{figure*}

\section{Discussion}
\label{sec:discussion}

\paragraph{OOD Detection} 
Our results show that exploiting the homophily of graph datasets using our GOOD method generally improves OOD detection on all GNNs and datasets.
Only for some datasets, like dblp-easy, GOOD is slightly outperformed by other combinations, \eg GCN and gDOC.
However, when looking into the \textit{overall best scores per dataset}, again we see that our GOOD method is on par with gDOC.
For example, on dblp-easy the best AUROC score is $0.68$ for gDOC on Graph-MLP, while GOOD on Graph-MLP reaches the same value.
The top performance of our GOOD method can be explained by the property of graph homophily.
For the homophile datasets Cora, CiteSeer, PubMed, and OGB-arXiv, our proposed method consistently improves the results.
For the heterophile datasets dblp-easy and dblp-hard, the improvement is lower or results even get worse by a small margin for some GNN models and OOD detectors, \eg GAT and ODIN on dblp-hard.
It shows that the homophily assumption of GOOD is necessary to be full-filled to guarantee improvement but does not diminish much if it is violated.
However, also for the heterophile datasets, there is always a combination of a GNN model and OOD detector that improves the results.
A promising combination of methods is Graph-MLP with IsoMax+, which performs best on Cora, Citeseer, and PubMed.
We assume that this is the effect of the Graph-MLPs use of a contrastive loss function during training that pulls the vertices in the $r$-hop neighborhood together, while the representation to other vertices is pushed away.
This allows IsoMax+ to learn good prototypes on the vertex representations for the classes, since it is also based on contrastive learning.
It will be interesting to study this synergy effect in future work. 
On dblp-hard, ODIN performed better, while gDOC is better on the other temporal datasets.
Notably, the accuracy can decrease using gDOC and IsoMax+ compared to ODIN (which is equal to a model without OOD detection).
gDOC reduces the accuracy on $6$ out of $12$ GNN and dataset combinations in the static setting and in $8$ out of $12$ in the temporal setting. 
IsoMax+ reduces the accuracy in $6$ out of $12$ in the static setting.
On the temporal datasets, IsoMax+ has the challenge that it needs to rearrange the prototypes in the embedding space when new classes appear.
For the new classes, new prototypes need to be introduced and the existing ones rearranged.

Please note that beside the $60\%$, $20\%$ and $20\%$ split used in the experiments of the static datasets, we also run an experiment with the Planetoid~\cite{Planetoid} split.
In that split, there are only $20$ vertices per class used for training, \ie it is a semi-supervised training setting.
From the results of the experiment with the Planetoid split, we come to the same conclusions.
The table of this experiment is provided in Appendix~\ref{sec:app_ssl}

\begin{figure*}[!ht]
    \centering
    \includegraphics[width=0.9\textwidth]{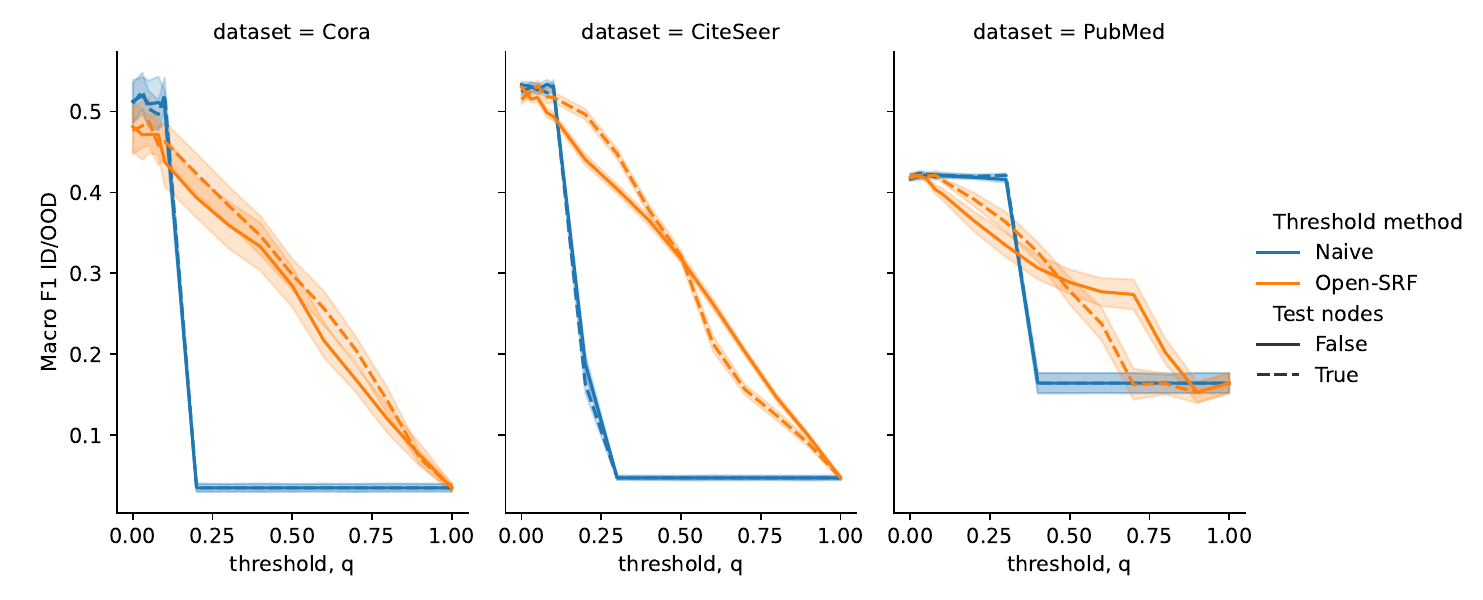}
    \caption{The results of the binary OOD vs. ID classification over different values of the OOD threshold (naive) and hyperparameter $q$ (our Open-WRF). 
     We used GCN and ODIN for all datasets, namely Cora (left), CiteSeer (middle), and PubMed (right).
     The dashed lines indicate that OOD scores of the test vertices are used.
     }
    \label{fig:res_threshold}
\end{figure*}

\paragraph{Detailed Analysis of GOOD's Neighborhood Parameter}

In order to assess the impact of the neighborhood influence parameter $\alpha_\mathrm{OOD}$ on the performance of the GOOD method, we systematically varied $\alpha_\mathrm{OOD}$ from $0.0$ to $1.0$ in steps of $0.1$, where $0.0$ corresponds to ignoring the neighborhood information and $1.0$ corresponds to ignoring the individual score. 
We conducted experiments for Graph-MLP and IsoMax+ using the static datasets following the procedure in Section~\ref{sec:experiment_app}.
The results were obtained by averaging over $10$ runs and are presented in Figure \ref{fig:neighinfluence_static_datasets_622}.
We observe that there is always an optimum for some $\alpha_\mathrm{OOD} \neq 0$, which means that using GOOD always improves the result with a favorable choice of the hyperparameter.
In general, the best values are between $0.2$ to $0.6$ on all datasets.
The largest improvement is on Cora, which is due to the high homophily of the dataset.
GOOD improves the score for every value on Cora except for the extreme case $\alpha_\mathrm{OOD}=1.0$. 
On CiteSeer and PubMed, GOOD loses some performance for large values of $\alpha_\mathrm{OOD}$.
However, the main results in Table~\ref{tab:main_results} show that an adequate value can be found by tuning $\alpha_\mathrm{OOD}$ on the validation set.

\paragraph{Open World Classification}
With the fixed threshold, Open-WRF is on par with the baseline methods. 
The baseline methods underperformed, since $\delta = 0.1$ or $q=0.1$ was a rather sub-optimal choice for the threshold.
This reflects a realistic setting, since the estimation of the ratio of OOD vertices can be quite inaccurate.
Our robust Open-WRF method compensates this sub-optimal threshold selection.
On dblp-easy and OGB-arXiv the scores are the same since these datasets contain many time steps and vertices without new classes, where even Open-WRF could not compensate for the rather high threshold.

\paragraph{Detailed Analysis of OOD Threshold Determination}

We further analyze the influence of the threshold parameter $q$ on the macro F1 score to validate the robustness of our Open-WRF method on the static datasets.
We select GCN as the most common GNN, together with ODIN, since it is a post-hoc method that can easily be applied to any trained model.
The results are provided in Figure~\ref{fig:res_threshold}.
The orange lines are our Open-WRF method, while the blue lines are naively applied hard thresholds $\delta$, as described in Section~\ref{sec:experiment_app}.
Solid lines represent scores that are purely computed on train vertices, and dashed lines represent scores that are computed including the test vertices.
The orange lines are almost always above the blue ones.
They also show a slower decrease.

This shows that our Open-WRF method with the hyperparameter $q$ is more robust in terms of determining whether datapoints are OOD than setting a naive (fixed) threshold.
This is due to the weakly-supervised learning procedure of Open-WRF, which only requires the hyperparameter $q$, a rough estimate of how many new OOD vertices are expected in the next time step.
Since the orange slopes decrease much slower, the range of good thresholds for OOD detection is increased when using Open-WRF compared to applying any fixed threshold. 
For PubMed, the F1 scores are considerably lower, which can be explained by the lower AUROC scores observed for PubMed (see Table~\ref{tab:main_results}).
Lower AUROC scores make it more difficult to separate the ID from OOD datapoints.

It can also be seen that the highest macro F1 scores are obtained by the naive method which tries out all possible values of the threshold.
For example, the threshold $0.2$ on CiteSeer produces F1 scores higher than those for both of the Open-WRF variants.
This is because there is always an optimal naive threshold, which is an upper bound to Open-WRF.
However, in practice, it is not feasible to determine this optimal threshold since there is no ground truth for OOD samples.
In general, the quality of the OOD scores poses an upper bound to the overall performance of Open-WRF, as it is the case for all threshold determination methods.
A strong advantage of our Open-WRF method is that the hyperparameter $q$ has an intuitive interpretation.
Since $q$ models the assumed percentage of OOD vertices in the next time step, it can be estimated from real-world datasets.

\paragraph{Assumptions and Limitations}
Our meta-method GOOD assumes homophily in the graph data.
While this could be considered a limitation, it is likewise also a very reasonable assumption.
The GNN models like those considered (GCN, GAT, GraphSAGE, and Graph-MLP) and others assume homophily and, thus, perform well when the graphs have high homophily scores.
For Open-WRF one needs to choose the parameter $q$.
However, choosing some kind of threshold is always necessary for practical applications that require a crisp decision.
In contrast to naive thresholds, $q$ can be interpreted as the expected ratio of OOD nodes and is less sensitive to small errors in the selection of the parameter.
Nevertheless, the value of $q$ is in general rather small, which leads to an imbalanced dataset for Open-WRF. 
This issue can be mitigated by employing standard imbalanced training techniques such as loss weighting or sampling strategies. 
We experimented with six datasets of different domains.
Three graphs are static and three temporal, including four homophile (Cora, CiteSeer, PubMed, and OGB-arXiv) and two heterophile (dblp-easy and dblp-hard) graphs to study the effect of violating the homophily assumption on GOOD.
Further studies with, \eg synthetic graphs may be conducted, where we explicitly control the homophily scores in the graph generation.
This way, we can investigate the influence of graph homophily on both the vertex classification performance and ability to detect OOD classes.
Furthermore, the hyperparameter tuning on the temporal graphs may be influenced by the growth of the graphs over time.
This is intended because it resembles the real-world challenges of lifelong graph learning.
The results are comparable, as we applied the same train/test procedure to every GNN model. 
In future work, one may consider re-adapting the hyperparameters after some sequence of $m$ many tasks in lifelong learning.

\section{Conclusion}\label{sec:conclusion}
We have proposed a new way to aggregate OOD scores in graph-structured data, GOOD, whose effectiveness was confirmed with four GNN models on three static and three temporal graph datasets. 
Our experiments show that GOOD can improve the OOD detection performance by a large margin, while it never decreases the performance. 
To transition from OOD scores to concrete decisions, \ie to decide if a datapoint is OOD or ID, we have revisited the problem of determining thresholds for OOD detection. Here, we have introduced a weakly-supervised relevance feedback method, Open-WRF, which substantially decreases the sensitivity to thresholds in OOD detection. 
Both GOOD and Open-WRF can be applied in conjunction, not as a replacement, with existing  methods for OOD detection.

\extended{Some future extensions to our work have been suggested above.
Furthermore, we are interested in evaluating the influence of window sizes, as proposed by \citet{Galke_LL}, on OOD performance with limited data access.}

\printbibliography

\FloatBarrier
\clearpage

\appendix

\subsection{Graph Neural Network Models}
\label{sec:appendix_models}

\paragraph{Graph Convolutional Network}
\citet{GCN} proposed the Graph Convolutional Network (GCN), where they introduced a message passing GNN on the symmetric normalized adjacency matrix.
The update equation for each layer is: 
\begin{equation}
    H^{l+1} = \sigma \left( \tilde{D}^{-\frac{1}{2}}\tilde{A}\tilde{D}^{-\frac{1}{2}}H^lW^l \right)
\end{equation}
where $\tilde{A} = A + I $ is the adjacency matrix with self-loops of the undirected graph, $\tilde{D}$ is the degree matrix of $\tilde{A}$, $W$ is a layer-specific trainable weight matrix, and $\sigma$ some non-linearity.
For the first layer, we have $H^0 = X$, \ie the GCN propagates and updates feature representations across the graph structure. 

\paragraph{Graph Attention Network}
The Graph Attention Network (GAT) was proposed by \citet{GAT}.
The architecture introduces an attention mechanism for graphs to weight edges in the message passing by a learned function. 
For each layer, GAT uses an attention mechanism $a$ to compute the attention coefficients for each edge,
\begin{equation}
    e_{ij} = a\left(Wh_i, W_j\right)
\end{equation}
 where $h_i$ is the representation of vertex $i$,  $h_j$ the representation of vertex $j$, and $W$ a shared learnable weight matrix.

These coefficients are normalized by a softmax function to
\begin{equation}
    a_{ij} = \softmax_j(e_{ij}) = \frac{\exp(e_{ij})}{\sum_{k \in \mathcal{N}(v_i)} \exp(e_{ik})}.
\end{equation}

Typically the attention mechanism $a$ is implemented by a single layer feedforward neural network. 
The final representation of a vertex $i$ is then computed by:

\begin{equation}
    h_i^{l+1} = \sigma \left( \sum_{j \in \mathcal{N}(v_i)} a_{ij}Wh^l_j \right)
\end{equation}

This mechanism is extended by multi-head attention, similarly to \cite{attention_all_you_need}, where multiple of the representations are kept at once and concatenated.
In the final layer, these representations are averaged for the final classification. 

\paragraph{GraphSAGE}
GraphSAGE has been proposed by \citet{graph-sage}. 
We use GraphSAGE-mean, where the messages are aggregated by computing their mean.
It samples a fixed number of training nodes (batch size) and adds neighbors for each sampled node up to a predefined neighborhood size.
For each selected neighbor, the procedure is repeated, \ie a fixed number of their neighbors is sampled.
The number of times this is repeated, \ie the number of hops of the sampler, is a hyperparameter. 
Note that if the neighborhood size is small, not all neighbors are considered at each training epoch, which results in different neighborhoods for a node at each epoch.
We used it with a batch size of $512$ on the static datasets and neighborhood sizes of $25$ for the first and $5$ for each further layer of the neural network, as in the work of \citet{graph-sage}.

On the temporal dataset, the batch size has been set to $20$ percent of the training graph for the current time step, 
expanded by $50$ neighbors per vertex for the first and $20$ neighbors per vertex for each subsequent layer.

\paragraph{Graph-MLP}
Graph-MLP, proposed by \cite{Graph-MLP}, is a model based on the common MLP that utilizes the graph structure in the loss function during the training process, given by the following equation:

\begin{equation}
    l_i = -\log\frac{\sum_{i=1}^B \mathbf{1}_{i \neq j} \gamma_{ij}\exp(sim(z_i, z_j)/\tau) }{\sum_{k=1}^B \mathbf{1}_{i \neq k}\exp(sim(z_i, z_k)/\tau)},
\end{equation}
where $sim$ is the cosine similarity, $B$ the batch size and $\tau$ a temperature parameter.
The factor $\gamma_{ij}$ is non-zero if and only if $j$ is in the $r$-hop neighborhood of $i$.
This loss is averaged over the uniformly sampled batch of vertices and combined with the cross entropy loss for vertex classification to:

\begin{equation}
    loss_{CE} + \beta \cdot loss_{NC},
\end{equation}
with the scaling parameter $\beta$.

The original Graph-MLP does not specify how the model can be extended to multiple layers.
In this work, we repeat the linear layer, activation function, and layer normalization structure $N-1$ times, where $N$ is the number of layers, and finish with a linear layer as in the original paper.
The contrastive loss is still applied to the penultimate linear layer.

After setting the parameters for Graph-MLP, we optimized the other model-specific hyperparameters separately over the range $\alpha \in \{0, 1, 10, 20, 100\}$ and $\tau \in \{0.1, 1, 2\}$ to see if further improvement is possible and evaluate how the loss weight relates to the graph homophily.

For validation of the re-implementation and comparison to the inductive setting, we also applied Graph-MLP in a transductive manner on the static datasets in a regular vertex-classification setting.
See Table~\ref{tab:graph-mlp_transductive} for the results.
They are the same as in the original GraphMLP paper of \citet{Graph-MLP}.

\begin{table}[!ht]
\centering
\normalsize		
    \begin{tabular}{cccc}
        \toprule
        Dataset & Cora & CiteSeer & PubMed \\
        \midrule
        Test accuracy & $0.778$ & $0.697$ & $0.776$ \\
        \bottomrule
    \end{tabular}
    \caption{ Test accuracy for our re-implemented version of Graph-MLP on the static datasets in a transductive setting.}
    \label{tab:graph-mlp_transductive}
\end{table}

\subsection{Details on Temporal Datasets}
\label{sec:ds_appendix}
\begin{figure*}[!th]
   \begin{subfigure}[b]{0.45\textwidth}
         \centering
         \includegraphics[width=\textwidth]{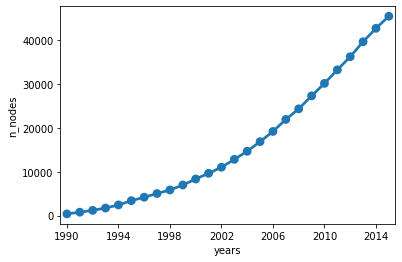}
        \caption{The cumulative number of vertices per year for the dblp-easy dataset.}
        \label{fig:dblp-easy_nodes}
     \end{subfigure}
     \begin{subfigure}[b]{0.425\textwidth}
         \centering
         \includegraphics[width=\textwidth]{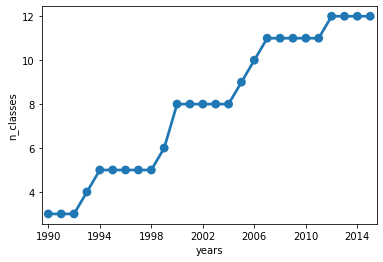}
         \caption{The number of known classes per year for the dblp-easy dataset.}
        \label{fig:dblp-easy_classes}
     \end{subfigure}
     \caption{Vertices and classes per year for the dblp-easy dataset.}
     \label{fig:characteristics_dblp-easy}
\end{figure*}

\begin{figure*}[!th]
   \begin{subfigure}[b]{0.45\textwidth}
         \centering
         \includegraphics[width=\textwidth]{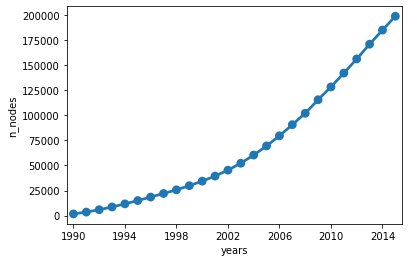}
           \caption{The cumulative number of vertices per year for the dblp-hard dataset.}
          \label{fig:dblp-hard_nodes}
     \end{subfigure}
     \begin{subfigure}[b]{0.425\textwidth}
         \centering
         \includegraphics[width=\textwidth]{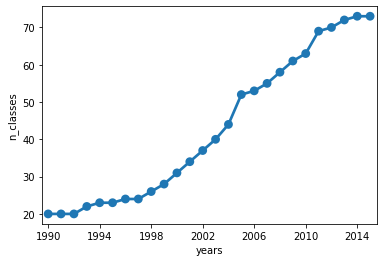}
         \caption{The number of known classes per year for the dblp-hard dataset.}
        \label{fig:dblp-hard_classes}
     \end{subfigure}
     \caption{Vertices and classes per year for the dblp-hard dataset.}
     \label{fig:characteristics_dblp-hard}
\end{figure*}

New classes are continually appearing in the dblp-easy and dblp-hard datasets over the years, as one can see in Figures~\ref{fig:characteristics_dblp-easy} and \ref{fig:characteristics_dblp-hard}.

\begin{figure*}[!th]
   \begin{subfigure}[b]{0.45\textwidth}
         \centering
         \includegraphics[width=\textwidth]{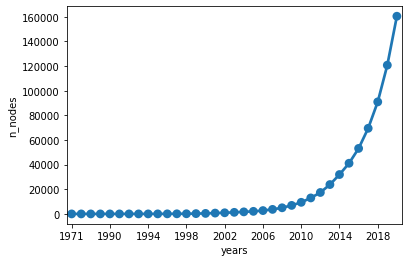}
         \caption{The cumulative number of vertices per year for the OGB-arXiv dataset.}
         \label{fig:arxiv_nodes}
     \end{subfigure}
     \begin{subfigure}[b]{0.425\textwidth}
         \centering
         \includegraphics[width=\textwidth]{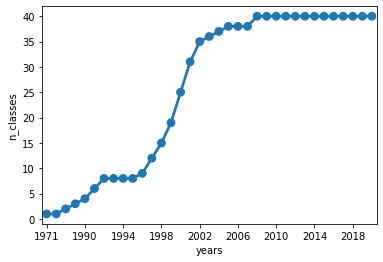}
        \caption{The number of known classes per year for the OGB-arXiv dataset.}
        \label{fig:arxiv_classes}
     \end{subfigure}
     \caption{Vertices and classes per year for the OGB-arXiv dataset.}
     \label{fig:characteristics_arxiv}
\end{figure*}

For OGB-arXiv, the number of train and validation vertices, \ie the number of vertices in time step $t_0$ and $t_{1}$ is very low.
This  is because the total number of $40$ classes saturates early ($2007$), while the amount of vertices is $4980$, which is only about $3\%$ of the nodes in the dataset.
The number classes and nodes over the years can be seen in Figure~\ref{fig:characteristics_arxiv}.

\subsection{Homophily}
\label{sec:homphily_appendix}
We conducted further analyses regarding the homophily and structure of the datasets.
In Table~\ref{tab:homophily_index_vs_inter_intra_class_edges} one can see the number of inter-class edges $C_I$, \ie the edges that connect vertices of different classes, the number of intra-class edges $C_E$, \ie the edges that connect vertices of the same class and the homophily index \citep{homophily_index}. 
The homophily index is given by: 
\begin{equation*}
\frac{|C_E| - |C_I|}{|E|}
\end{equation*}
where $|E|$ represents all edges.
This index is between $-1$ and $1$.
The index is $-1$, if the graph is completely homophile and $1$ if it is completely heterophile.

\begin{table}[!th]
    \centering
    \small
    \begin{tabular}{lrrr}
         \toprule
         Dataset  & Inter-class & Intra-class & Homophily-index\\
         \toprule
        Cora        & $2,006$     & $8,550$  &   $-0.610$ \\
        CiteSeer    & $2,408$    &  $6,696$ & $-0.471$ \\
        PubMed      & $17,518$    & $71,130$  &  $-0.605$ \\
        \midrule
        dblp-easy  & $75,543$ & $191,684$ & $0.435$ \\
        dblp-hard  & $238,765$ & $1,241,412$ & $0.677$ \\
        OGB-arXiv  & $763,986$ & $402,257$ & $-0.310$ \\
        \bottomrule
    \end{tabular}
    \caption{Homophily scores of the inter-class and intra-class edges versus the homophily index.}
    \label{tab:homophily_index_vs_inter_intra_class_edges}
\end{table}

Since we perform a leave-one-class-out procedure on the static datasets, we computed the homophily index per class as presented in Table~\ref{tab:per_class_graph_level_homophily_static}.
It can be observed that the homophily of separate classes does not differ much in the same dataset.

\begin{table*}[!th]
    \centering
    \begin{tabular}{llrrrrrrr|rr}
         \toprule
         Dataset  & \# classes & $0$ & $1$ & $2$ & $3$ & $4$ & $5$ & $6$ & Avg. & Std.\\
         \midrule
        Cora   & $7$ &$-0.708$ & $-0.693$ & $-0.678$ & $-0.688$ & $-0.688$ & $-0.697$ & $-0.697$ & $-0.692$ & $0.008$\\
        CiteSeer    & $6$ &$-0.764$ & $-0.674$ & $-0.630$ & $-0.644$& $-0.637$ & $-0.638$ & - & $-0.666$ & $0.046$\\
        PubMed      & $3$ &$-0.341$ & $-0.147$ & $-0.183$ & - & - & - & - & $-0.223$ & $0.084$\\
        \bottomrule
    \end{tabular}
    \caption{The homophily index for each class for the static datasets along with the average over the classes and the standard deviation.}
    \label{tab:per_class_graph_level_homophily_static}
\end{table*}

We considered three further homophily measures, based on graph-level, vertex-level, and class-insensitive edge homophily.
All give a score between $0$ and $1$, where $1$ is completely homophile.
The graph-level homophily ratio~\citep{edge-homophily-ratio} is defined as
\begin{equation}
    \frac{|\{(v,w) : (v,w) \in E \text{ and } y_v = y_w \}|}{|E|}\,.
\end{equation}

The vertex-level homophily ratio~\citep{node-homophily-ratio} is defined as
\begin{equation}
    \frac{1}{|V|}\sum_{v \in V} \frac{|(w,v): w \in \mathcal{N}(v) \text{ and } y_v=y_w|}{\mathcal{N}(v)}\,.
\end{equation}

Furthermore, we consider the class-insensitive edge homophily ratio~\citep{class-insensitive-edge-homophily-ratio}.
This measure computes the homophily per class and a computes the unweighted average over the classes.
This unweighted class-based homophily score is defined as 
\begin{equation}
    \frac{1}{|C|-1} \sum^{|C|}_{k=1} \max \left(0, h_k - \frac{|C_k|}{|V|} \right)\,,
\end{equation}

with $h_k$ being the average class-based homophily of the graph, \ie the ratio of inter-class edges to all edges of class $k$, and $C$ the set of possible classes.
These are the homophily measures used in Table~\ref{tab:homohily_static}.

\begin{figure*}
     \centering
     \begin{subfigure}[b]{0.3\textwidth}
         \centering
         \includegraphics[width=\textwidth]{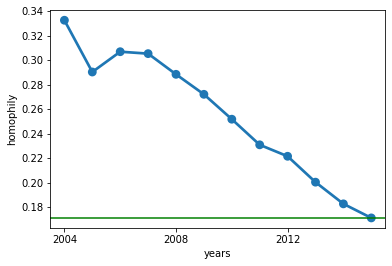}
         \caption{Class insensitive edge homophily per time step on dblp-easy.}
    \label{fig:dblp-eay_cumulative_homophily}
     \end{subfigure}
     \hfill
     \begin{subfigure}[b]{0.3\textwidth}
         \centering
         \includegraphics[width=\textwidth]{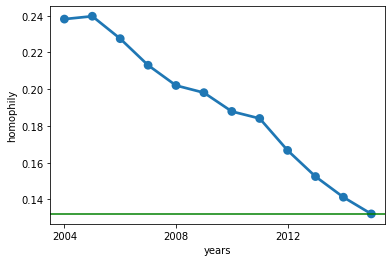}
         \caption{Class insensitive edge homophily per time step on dblp-hard.}
    \label{fig:dblp-hard_cumulative_homophily}
     \end{subfigure}
     \hfill
     \begin{subfigure}[b]{0.3\textwidth}
         \centering
         \includegraphics[width=\textwidth]{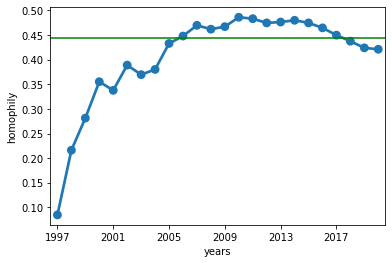}
          \caption{Class insensitive edge homophily per time step on OGB-arXiv.}
    \label{fig:ogb-argiv_cumulative_homophily}
     \end{subfigure}
      \caption{Cumulative class insensitive edge homophily per time step on each of the temporal datasets. The solid line (green) represents the homophily on the whole graph.}

      \label{fig:temporal_homophily}
\end{figure*}

On the temporal datasets, the class-insensitive edge homophily has been computed per time step.
For the dblp datasets, the homophily is comparably low and is falling towards the homophily of the whole graph, as show in Figures~\ref{fig:dblp-eay_cumulative_homophily} and \ref{fig:dblp-hard_cumulative_homophily}.
For OGB-arXiv, the homophily exceeds the homophily of the whole graph in the last third of the years, as shown in Figure~\ref{fig:ogb-argiv_cumulative_homophily}.
This is because OGB-arXiv is a common citation graph with research topics as classes.
Here, the vertices are papers that mostly cite other papers in their own research area.
In contrast, the dblp datasets are citation graphs, with conferences and journals as classes.
Since authors cite papers from many different conferences and journals, the homophily drops as the number of classes grows.

\subsection{Hyperparameters}
\label{sec:hp_appendix}

For each of the static datasets, we tuned the models on the following parameters via grid search: 

\begin{itemize}
    \item Number of layers $L \in \{2, 3\}$
    \item Hidden dimension $h \in \{32, 64, 128, 256, 512\}$
    \item Dropout rate $r_d \in \{ 0.4, 0.5, 0.6, 0.7, 0.8,0.9 \}$
    \item Learning rate $l \in \{0.0001, 0.001, 0.01, 0.1 \}$
    \item Training epochs: $e \in \{100, 200, 300, 500\}$
\end{itemize}

Some model-specific parameters were included in the grid search as well. 
For GAT this is the number of attention heads $ht \in \{4, 8, 16, 32\}$ and the attention dropout rate $r_a \in \{0.2, 0.3, 0.4, 0.5\}$.
For Graph-MLP, we evaluated different values for the power of the adjacency matrix $r \in \{1,2,3\}$.
The other model-specific parameters were set to default values, \ie the weighting of the specialized contrastive loss $\alpha=1$ and the temperature parameter $\tau = 1$, to keep the grid parameter search feasible. 

For the temporal datasets, we tried different parameters, \ie larger hidden dimension and number of layers, since the datasets consist of more datapoints and classes:

\begin{itemize}
    \item Number of layers $L \in \{2, 3,4\}$
    \item Hidden dimension $h \in \{128, 256, 512, 1024, 2048\}$
    \item Training epochs: $E \in \{100, 200, 300\}$
\end{itemize}

For GAT and Graph-MLP, additional parameters have been tested.
We tried hidden dimensions in $\{8, 16, 32, 64, 128\}$ and number of attention heads in $\{2,4,8\}$ for GAT and for Graph-MLP adjacency matrix powers $r \in \{1,2,3\}$.

The ODIN parameters were tuned on
\begin{itemize}
    \item $T \in \{1,10,100,1000\}$
    \item $\eps \in \{0.01,0.05, 0.08, 0.1, 0.2, 0.5, 0.8, 0.9 \}$
\end{itemize}.

As an optimizer, Adam \citep{adam} was used with zero weight decay to stay in line with the theoretical motivation of Adam, as stated by \citet{NoWeightDecayAdam}.
The authors showed that due to some mistakes in the formal derivation of the weight decay for Adam, it is not equivalent to $L_2$ regularization as usual. 
Therefore, there is no theoretical foundation in using weight decay with Adam. 
Each experiment has been repeated $10$ times, to increase the validation of the hyperparameters.
We choose the hyperparameters the models performed best on average over the $10$ runs.

The final hyperparameter values can be found in the following Tables~\ref{tab:gcn_hp} to \ref{tab:graph-mlp_hp}.

\begin{table*}[!th]
    \centering
    \begin{tabular}{lrrrrr}
        \toprule 
         Dataset & Layer & Hidden dimension & Dropout rate & Learning rate &  Epochs  \\
         \toprule
         Identity  &&&&& \\
         \midrule
         Cora & $2$ & $128$ & $0.8$ & $0.001$ & $200$  \\
         CiteSeer & $2$ & $256$ & $0.8$ & $0.01$ & $200$ \\	
         PubMed & $2$ & $128$ & $0.8$ &$ 0.01$ & $300$ \\
         dblp-easy & $2$ & $2,048$ & $0.9$ & $0.01$  &  $200$  \\
         dblp-hard  &$2$& $2,048$  & $0.9$ & $0.001$ & $200$ \\
         OGB-arXiv & $2$ & $1,048$ & $0.6$ & $0.01$  &  $200$  \\
         \midrule
         IsoMax+ &&&&& \\
         \midrule
        Cora & $3$ & $128$ & $0.7$ & $0.001$ & $300$  \\
        CiteSeer & $2$ & $128$ & $0.9$ & $0.001$ & $300$ \\	
        PubMed & $2$ & $256$ & $0.8$ & $0.001$ & $300$ \\
         dblp-easy & $3$ & $2,048$ & $0.8$ & $0.1$  &  $100$  \\
         OGB-arXiv &$2$ &$2,048$ & $0.8$ & $0.1$ & $300$ \\
         \midrule
         gDOC &&&&& \\
         \midrule
         Cora &  $2$  &  $128$  &  $0.6$ & $0.001$ & $300$	\\
          CiteSeer & $2$ & $64$ & $0.8$ & $0.001$ & $300$ \\
          PubMed & $2$ & $256$ & $0.7$ & $0.001$ & $300$ \\
          dblp-easy & $2$ & $2,048$ & $0.9$ & $0.0001$ & $300$ \\
          dblp-hard & $2$ & $2,048$ & $0.6$ & $0.0001$ & $300$ \\
          OGB-arXiv & $2$ & $1,024$ & $0.7$ & $0.001$ & $200$ \\	
         \bottomrule
    \end{tabular}
    \caption{Best hyperparameter values for GCN on each dataset and OOD detector.}
    \label{tab:gcn_hp}
\end{table*}

\begin{table*}[!th]
    \centering
    \begin{tabular}{lrrrrr}
        \toprule 
         Dataset & Layer & Hidden dimension & Dropout rate & Learning rate &  Epochs  \\
         \toprule
         Identity  &&&&& \\
         \midrule
        
          Cora & $2$ & $256$ & $0.7$ & $0.001$ & $300$  \\
         CiteSeer & $2$ & $256$ & $0.8$ & $0.001$ &  $200$\\
         Pubmed &$2$& $128$  & $0.9$ & $0.001$ &  $200$ \\
         \midrule
         IsoMax+ &&&&& \\
         \midrule
        Cora & $3$ & $128$ & $0.7$ & $0.001$ & $300$  \\
        CiteSeer & $2$ & $128$ & $0.9$ & $0.001$ & $300$ \\	
        PubMed & $2$ & $256$ & $0.8$ & $0.001$ & $300$ \\
         Cora & $2$ & $256$ & $0.9$ & $0.01$ & $300$  \\
         CiteSeer & $2$ & $256$ & $0.9$ & $0.1$ &  $300$  \\
         Pubmed &$2$& $64$  & $0.7$ & $0.01$ & $300$ \\
         \midrule
         gDOC &&&&& \\
        \midrule
     
          Cora & $2$ &	$256$	 & $0.8$	& $0.0001$ & $300$ \\
          CiteSeer & $2$ & $256$ & $0.9$ & $0.0001$ &  $300$\\
          Pubmed  &$3$& $64$  & $0.8$ & $0.001$ &  $200$ \\
         \midrule
         \bottomrule
    \end{tabular}
    \caption{Best hyperparameter values for GCN on each dataset and OOD detector on the Planetoid split.}
    \label{tab:gcn_hp}
\end{table*}

\begin{table*}[!th]
    \centering
    \begin{tabular}{lrrrrrrr}
        \toprule 
         Dataset & Layer & Hidden dimension & Dropout rate & Learning rate & Sample size & Epochs  \\
         \midrule
         Identity &&&&& \\
         \midrule
         Cora & $2$ & $256$ & $0.9$ & $0.001$ & $128$ &$100$ \\
         CiteSeer & $2$ & $256$ & $0.9$ & $0.0001$ & $128$ &$300$ \\
         PubMed & $2$ & $256$ & $0.9$ & $0.001$ & $128$ &$300$	\\
          Cora & $3$ & $256$ & $0.8$ & $0.001$ & $32$ &  $100$ \\
         CiteSeer & $2$ & $256$ & $0.8$ & $0.0001$ & $32$ & $100$ \\
         PubMed & $2$ & $256$ & $0.6$ & $0.001$ & $32$ & $200$ \\
         dblp-easy & $3$ & $2,048$ & $0.8$ & $0.001$ & - &$300$  \\
         dblp-hard & $2$	& $2,048$   & $0.8$	 & $0.0001$ & - & $300$ \\
         OGB-arXiv &  $2$& $2,048$ &$0.9$	 &$0.001$& - & $200$ \\
         \midrule
        IsoMax+ &&&&& \\
        \midrule
   
          Cora & $2$ & $256$ & $0.7$ & $0.01$ & $32$ & $100$ \\
        CiteSeer & $2$ & $128$ & $0.6$ & $0.01$ & $32$ & $100$ \\
        PubMed &  $2$ & $256$ & $0.7$ & $0.01$ & $32$ & $200$ \\
        dblp-easy & $2$ & $1,024$ & $0.6$ & $0.001$ & $200$ \\
        OGB-arXiv & $2$ & $512$ & $0.6$ & $0.01$ & - & $300$ \\
         \midrule
        gDOC &&&&& \\
        \midrule
        Cora & $2$ & $128$ & $0.8$ & $0.0001$ & $32$ &$300$ \\
        CiteSeer & $2$ & $256$ & $0.7$ & $0.0001$ & $32$ & $200$ \\
        PubMed & $2$ & $256$ & $0.8$ & $0.01$ & $32$ & $100$ \\	
         dblp-easy & $2$ & $2,048$ & $0.9$ & $0.0001$ & - & $300$ \\
         dblp-hard & $3$ & $2,048$ & $0.8$ & $0.001$ & - & $200$ \\
         OGB-arXiv & $3$ & $1,024$ & $0.8$ & $0.0001$ & - & $200$ \\	
         \bottomrule
    \end{tabular}
    \caption{Best hyperparameter values for GraphSage on each dataset and OOD detector. The batch size on the temporal dataset was 20\% of the training data for the current time step.}
    \label{tab:sage_hp}
\end{table*}

\begin{table*}[!th]
    \centering
    \begin{tabular}{lrrrrrrr}
        \toprule 
         Dataset & Layer & Hidden dimension & Dropout rate & Learning rate & Sample size & Epochs  \\
         \midrule
         Identity &&&&& \\
         \midrule
        Cora & $2$ & $256$ & $0.9$ & $0.001$ & $128$ &$100$ \\
         CiteSeer & $2$ & $256$ & $0.9$ & $0.0001$ & $128$ &$300$ \\
         PubMed & $2$ & $256$ & $0.9$ & $0.001$ & $128$ &$300$	\\
         \midrule
        IsoMax+ &&&&& \\
        \midrule
        Cora & $2$ & $128$ & $0.8$ & $0.01$ & $128$ &$300$ \\
        CiteSeer & $2$ & $64$ & $0.6$ & $0.01$ & $128$ & $300$ \\
        PubMed & $2$ & $64$ & $0.6$ & $0.01$ & $128$ & $300$\\
         \midrule
        gDOC &&&&& \\
        \midrule
        Cora & $2$ & $256$ & $0.6$ & $0.0001$ & $128$ &$300$  \\
         CiteSeer & $2$ & $256$ & $0.8$ & $0.001$ & $128$  &$300$  \\
         Pubmed &$2$& $256$  & $0.9$ & $0.001$ & $128$ & $300$ \\

\bottomrule
    \end{tabular}
    \caption{Best hyperparameter values for GraphSage on each dataset and OOD detector on the Planetoid split}
    \label{tab:sage_hp}
\end{table*}

\begin{table*}[!th]
    \centering
    \begin{tabular}{lrrrrrrr}
        \toprule
      
         Dataset & Layer& Hidden per head & Attn heads & Attn dropout & Dropout rate & Learning rate & Epochs  \\
         \toprule
         Identity &&&&&&& \\
         \midrule
        Cora & $3$ & $32$ & $16$ & $0.5$ & $0.9$ & $0.001$ & $200$\\
         CiteSeer & $3$ & $32$ & $8$ & $0.2$ & $0.7$ & $0.0001$ & $300$ \\
         PubMed & $3$ & $16$ & $16$ & $0.3$ & $0.8$ & $0.001$ & $300$ \\
         dblp-easy & $3$ & $128$ & $8$ & $0.3$ & $0.9$ &$0.001$ & $200$  \\
         dblp-hard & $3$ & $128$ & $8$ &$0.2$ & $0.9$ & $0.001$ & $300$  \\
         OGB-arXiv &$3$& $128$  & $4$ & $0.2$ & $0.6$ & $0.01$ & $300$ \\
        \midrule
         IsoMax+ &&&&&&& \\
        \midrule
    
         Cora & $2$ & $8$ & $32$ & $0.4$ & $0.9$ & $0.001$ & $200$ \\
         CiteSeer & $2$ & $16$ & $32$ & $0.3$ & $0.9$ & $0.0001$ & $300$ \\	
         PubMed & $3$ & $32$ & $32$ & $0.2$ & $0.7$ & $0.001$ & $300$ \\	
         dblp-easy & $3$ & $32$ & $4$ & $0.2$ & $0.6$ &$0.01$ & $300$  \\
         OGB-arXiv & $3$ & $32$ & $16$ & $0.2$ & $0.6$ & $0.001$ & $300$ \\
         \midrule
         gDOC &&&&&&& \\
         \midrule         
          Cora & $3$ & $16$ & $16$ & $0.2$ & $0.8$ & $0.001$ & $200$ \\
          CiteSeer & $3$ & $32$ & 8 & $0.3$ & $0.8$ & $0.0001$ & $200$ \\
          PubMed & $2$ & $32$ & $32$ & $0.2$ & $0.8$ & $0.001$ & $300$ \\      
         dblp-easy & $3$ & $128$ & $2$ & $0.2$ & $0.9$ &$0.001$ & $200$  \\
         dblp-hard & $3$  & $128$ & $8$ & $0.3$ & $0.7$ & $0.001$ & $200$ \\
         OGB-arXiv & $3$ & $32$	& $8$ & $0.2$ & $0.8$ & $0.001$ & $200$ \\

         \bottomrule
    \end{tabular}
    \caption{Best hyperparameter values for GAT on each dataset and OOD detector.}
    \label{tab:gat_hp}
\end{table*}

\begin{table*}[!th]
    \centering
    \begin{tabular}{lrrrrrrr}
        \toprule
      
         Dataset & Layer& Hidden per head & Attn heads & Attn dropout & Dropout rate & Learning rate & Epochs  \\
         \toprule
         Identity &&&&&&& \\
         \midrule    
         Cora & $3$ & $8$ & $32$ & $0.5$ & $0.9$ &$0.001$ & $300$  \\
         CiteSeer & $2$ & $16$ & $8$ &$0.2$ & $0.7$ & $0.001$ & $300$  \\
         Pubmed &$2$& $32$  & $8$ & $0.2$ & $0.9$ & $0.001$ & $200$ \\
        \midrule
         IsoMax+ &&&&&&& \\
         \midrule
          Cora & $2$ & $8$ & $16$ & $0.4$ & $0.8$ &$0.01$ & $300$  \\
         CiteSeer & $3$ & $8$ & $8$ &$0.4$ & $0.6$ & $0.01$ & $300$  \\
         PubMed &$2$& $32$  & $16$ & $0.3$ & $0.9$ & $0.01$ & $300$ \\
         \midrule
         gDOC &&&&&&& \\
         \midrule
         Cora & $2$ & $16$ & $16$ & $0.4$ & $0.7$ &$0.0001$ & $300$  \\
         CiteSeer &  $2$ &	$32$ & $16$ & $0.5$ & $0.9$ & $0.0001$ & $200$\\
         PubMed   &$2$	  & $32$ &$8$	&  $0.2$&  $0.9$&  $0.001$ & $200$ \\
         \midrule

\bottomrule
    \end{tabular}
    \caption{Best hyperparameter values for GAT on each dataset and OOD detector on the Planetoid split.}
    \label{tab:gat_hp}
\end{table*}

\begin{table*}[!th]
    \centering
    \begin{tabular}{lrrrrrr}
        \toprule 
         Dataset & Layer& Hidden dimension & Dropout rate & Learning rate & r  & Epochs  \\
         \toprule
         Identity &&&&&& \\
         \midrule
         Cora & $3$  & $256$ & $0.7$ & $0.01$ & $2$  & $500$ \\
         CiteSeer & $3$ & $512$ & $0.7$  & $0.01$ & $3$ & $100$ \\
         PubMed & $3$ & $128$ & $0.4$ & $0.001$ & $2$  & $300$ \\
         dblp-easy & $2$	& $256$ & $0.8$ & $0.0001$ & $3$&$200$ \\
         dblp-hard & $4$	& $1,024$ & $0.4$ & $0.0001$ & $2$ & $300$ \\
         OGB-arXiv & $3$ & $1,024$ & $0.4$ & $0.0001$ & $2$ & $200$ \\
         \midrule
         IsoMax+ &&&&&& \\
         \midrule
 
         Cora & $3$ & $256$ & $0.6$ & $0.01$ & $3$ & $200$ \\
         CiteSeer & $2$ & $256$ & $0.7$ & $0.01$ & $3$ & $100$ \\
         PubMed & $3$ & $512$ & $0.3$ &  $0.0001$ & $2$ & $500$ \\
         dblp-easy & $4$	& $1,024$ & $0.8$ & $0.1$ & $2$&$300$ \\
         OGB-arXiv &$3$  &$512$ &$0.6$ &$0.1$ & $3$ & $300$ \\
         \midrule
         gDOC &&&&&& \\
         \midrule

         Cora & $3$ & $128$ & $0.7$ & $0.01$ & $2$ & $200$ \\
         CiteSeer & $3$ & $128$ & $0.3$ & $0.001$ & $2$ & $200$ \\
         PubMed & $3$ & $512$ & $0.3$ &  $0.0001$ & $2$ & $500$ \\
         dblp-easy & $2$ & $128$ & $0.9$ & $0.0001$ & $2$ & $200$ \\
         dblp-hard & $3$ & $2,048$ & $0.7$ & $0.0001$ & $2$ & $200$ \\
         OGB-arXiv & $3$ & $1,024$ & $0.7$ & $0.0001$ & $2$ & $200$ \\
         
         \midrule
         \bottomrule
    \end{tabular}
    \caption{Best hyperparameter values for Graph-MLP on each dataset and OOD detector.}
    \label{tab:graph-mlp_hp}
\end{table*}

\begin{table*}[!th]
    \centering
    \begin{tabular}{lrrrrrr}
        \toprule 
         Dataset & Layer& Hidden dimension & Dropout rate & Learning rate & r  & Epochs  \\
         \toprule
         Identity &&&&&& \\
         \midrule
         Cora & $3$ & $512$ & $0.8$ & $0.0001$ & $2$ & $200$  \\
         CiteSeer & $3$ & $512$ & $0.8$ & $0.001$ & $2$ & $100$  \\
         Pubmed &$3$& $512$  & $0.7$ & $0.0001$ & $3$ & $200$ \\

         \midrule
         IsoMax+ &&&&&& \\
         \midrule
         Cora & $2$ & $128$ & $0.7$ & $0.0001$ & $3$ & $300$  \\
         CiteSeer & $2$ & $128$ & $0.9$ & $0.001$ & $3$ & $300$  \\
         PubMed &$3$& $256$  & $0.6$ & $0.0001$ & $2$ & $100$ \\        

\midrule
         gDOC &&&&&& \\
         \midrule
         Cora      & $2$& $512$ &	$0.8$	& $0.0001$  & $2$ &  $100$\\
         CiteSeer & $3$ & $512$ & $0.8$  & $0.0001$ & $2$ & $100$	\\
         PubMed   & $2$	& $512$ &	$0.8$ &	$0.01$ & $2$ &	$100$ \\
         \midrule
        
         \bottomrule
    \end{tabular}
    \caption{Best hyperparameter values for Graph-MLP on each dataset and OOD detector on the Planetoid split}
    \label{tab:graph-mlp_hp}
\end{table*}

\begin{table*}[!th]
    \centering
    \begin{tabular}{lrrrrrr}
        \toprule
         & Cora & CiteSeer & PubMed & dblp-easy & dblp-hard & OGB-arXiv  \\
         \midrule
         GCN    &  $0.08$/$1000$ & $0.05$/$10$ & $0.01$/$10$ &$0.08$/$10$& $0.05$/$10$& $0.02$/$10$ \\
         GAT    &  $0.01$/$100$  & $0.01$/$100$&$0.01$/$1000$& $0.05$/$1000$ & $0.01$/$1000$	& $0.5$/$1$	\\
         GraphSAGE & $0.2$/$100$ & $0.01$/$1000$& $0.1$/$1$ & $0.01$/$1000$& $0.8$/$1000$ &$0.9$/$100$\\
         Graph-MLP &$0.01$/$100$& $0.01$/$1000$& $0.01$/$10$& $0.01/1000$& $0.01$/$1000$ & $0.01$/$1$\\
         \bottomrule
         
    \end{tabular}
    \caption{Best hyperparameter values for ODIN on each model and dataset. Left of ``/'' is the perturbation magnitude $\eps$, right is the temperature $T$. }
    \label{tab:my_label}
\end{table*}

\begin{table*}[!th]
    \centering
    \begin{tabular}{lrrr}
        \toprule
         & Cora & CiteSeer & PubMed  \\
         \midrule
         GCN    &  $0.05/1000$ & $0.1/100$ & $0.05/1000$\\
         GAT    &  $0.01/1000$ & $0.01/1000$ & $0.01/1000$ 	\\
         GraphSAGE & $0.01/1$ &  $0.08/1000$ & $0.01/1000$ \\
         Graph-MLP & $0.01/1000$ & $0.01/1000$ & $0.08/100$\\
         \bottomrule
         
    \end{tabular}
    \caption{Best hyperparameter values for ODIN on each model for the 622 datasets. Left of ``/'' is the perturbation magnitude $\eps$, right is the temperature $T$. }
    \label{tab:my_label}
\end{table*}

\begin{table*}[!th]
    \centering
    \begin{tabular}{llrr}
        \toprule 
         Dataset & Model & OOD-Method & $\alpha_\mathrm{OOD}$  \\
         \midrule
         Semi-supervised &&& \\
         \midrule
         Cora & GCN & ODIN & $0.2$ \\
         Cora & GAT & ODIN & $0.8$ \\
         Cora & GraphSage & ODIN & $0.9$ \\
         Cora & Graph-MLP & IsoMax+  &$0.8$ \\
         CiteSeer & GCN & ODIN & $0.5$ \\
         CiteSeer & GAT & ODIN & $0.6$ \\
         CiteSeer & GraphSage & gDOC & $0.7$ \\
         CiteSeer & Graph-MLP & IsoMax+ & $0.9$ \\
         PubMed & GCN & ODIN & $0.5$ \\
         PubMed & GAT & gDOC & $0.6$ \\
         PubMed & GraphSage & gDOC  & $0.5$ \\
         PubMed & Graph-MLP & IsoMax+ & $0.6$	\\
         \midrule
          Supervised &&& \\
         \midrule
         Cora & GCN & gDOC & $0.6$ \\
         Cora & GAT & ODIN & $0.4$ \\
         Cora & GraphSage & gDOC & $0.3$ \\
         Cora & Graph-MLP &  IsoMax+ &$0.4$ \\
         CiteSeer & GCN & ODIN & $0.8$ \\
         CiteSeer & GAT & ODIN & $0.2$ \\
         CiteSeer & GraphSage & gDOC & $0.8$ \\
         CiteSeer & Graph-MLP & IsoMax+ & $0.2$ \\
         PubMed & GCN & ODIN & $0.9$ \\
         PubMed & GAT & IsoMax+ & $0.6$ \\
         PubMed & GraphSage & gDOC  & $0.8$ \\
         PubMed & Graph-MLP & IsoMax+ & $0.4$	\\
         \midrule
         Temporal &&& \\
         \midrule
         dblp-easy & GCN & gDOC & $0.7$ \\
         dblp-easy & GAT & gDOC & $0.3$ \\
         dblp-easy & GraphSage & gDOC & $0.1$ \\
         dblp-easy & Graph-MLP &gDOC & $0.1$ \\ 
         dblp-hard & GCN & ODIN & $0.6$ \\
         dblp-hard & GAT & ODIN & $0.1$ \\
         dblp-hard & GraphMLP & gDOC & $0.5$ \\	
         OGB-arXiv & GCN & gDOC & $0.9$ \\
         OGB-arXiv & GraphSage & gDOC & $0.3$ \\
         OGB-arXiv & GAT & gDOC & $0.6$ \\
        OGB-arXiv & Graph-MLP & gDOC & $0.9$ \\
     
        \bottomrule
    \end{tabular}
    \caption{Neighborhood influence per model and method, determined on the validation set.}
    \label{tab:aggreg_hp}
\end{table*}

\subsection{OOD Aggregation}

\begin{figure*}
     \centering
     \begin{subfigure}[b]{0.3\textwidth}
         \centering
        \includegraphics[width = \textwidth]{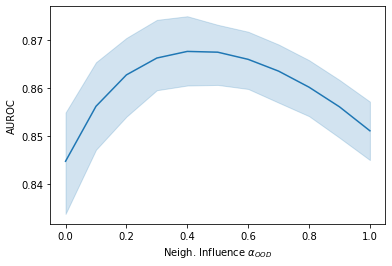}
        \caption{Cora}
        \label{fig:cora_neigh_inf}
     \end{subfigure}
     \hfill
     \begin{subfigure}[b]{0.3\textwidth}
         \centering
         \includegraphics[width =\textwidth]{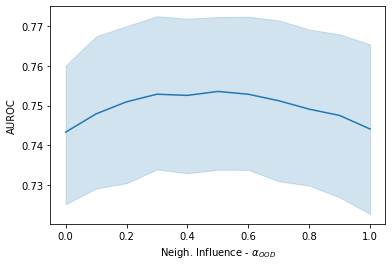}
        \caption{CiteSeer}
        \label{fig:citeseer_neigh_inf}
     \end{subfigure}
     \hfill
     \begin{subfigure}[b]{0.3\textwidth}
         \centering
         \includegraphics[width =\textwidth]{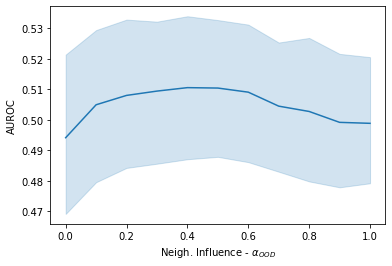}
    \caption{PubMed}
    \label{fig:pubmed_neigh_inf}
     \end{subfigure}
     \caption{Measurement of the influence of the values $\alpha_\mathrm{OOD}$ and the respective AUROC scores with confidence interval for the static datasets.}
    \label{fig:neighinfluence_static_datasets}
\end{figure*}

To analyze the function of the hyperparameter $\alpha_\mathrm{OOD}$ in the score aggregation, we swiped over $\alpha_\mathrm{OOD}$ values from $0.0$ to $1.0$ in steps of $0.1$ for datasets with different homophily scores. 
The result for the static datasets can be seen in Figure~\ref{fig:neighinfluence_static_datasets}.

\subsection{Semi-Supervised Setting}
\label{sec:app_ssl}
We use the Planetoid split by~\citet{Planetoid} as train, validation, and test split in an inductive setting to simulate temporal-semi-supervised node classification. 
The results are provided in Table~\ref{tab:semi-supervised}.

\begin{table}[!th]
\centering
\small
\begin{tabular}{lrrrr}
     \toprule
       &GCN      &GraphSAGE     &GAT        &Graph-MLP  \\
      \midrule
      \textbf{Static} & Acc/AUROC &  Acc/AUROC &  Acc/AUROC &  Acc/AUROC \\

\midrule
      Cora&&&&\\
     \midrule
     ODIN&  $0.820$/\underline{$0.845$} & $0.812$/\underline{$0.855$}  & $0.827$/\underline{$0.862$}  &$0.513$/\underline{$0.609$}\\
    IsoMax+& $0.794$/$0.719$  & $0.781$/$0.677$  & $0.791$/$0.796$  &$0.116$/$0.505$\\
     gDOC&  $0.659$/$0.723$ & $0.799$/$0.841$  & $0.801$/$0.843$  &$0.434$/$0.535$\\
     GOOD (own) & $0.820$/$\mathbf{0.863}$  & $0.812$/$\mathbf{0.877}$  & $0.827$/$\mathbf{0.883}$  &$0.510$/$\mathbf{0.742}$\\
     \midrule
      CiteSeer&&&& \\
     \midrule
     ODIN& $0.693$/\underline{$0.743$} & $0.696$/$0.715$   & $0.701$/\underline{$0.742$} & $0.352$/\underline{$0.551$}\\
     IsoMax+&  $0.648$/$0.644$ & $0.655$/$0.613$  & $0.640$/$0.683$  & $0.123$/$0.510$\\
     gDOC& $0.381$/$0.496$ & $0.701$/\underline{$0.717$}  & $0.685$/$0.701$  & $0.473$/$0.550$\\
    GOOD (own) & $0.693$/$\mathbf{0.753}$  & $0.688$/$\mathbf{0.773}$  & $0.700$/$\mathbf{0.749}$  &$0.349$/$\mathbf{0.621}$\\
     \midrule
     PubMed&&&& \\
     \midrule
     ODIN& $0.867$/$0.488$  & $0.862$/$0.546$  &$0.867$/$0.524$   & $0.738$/$0.497$\\
     IsoMax+& $0.765$/$0.505$  &$0.850$/$0.510$  & $0.855$/$0.539$  & $0.272$/$\underline{0.602}$\\
     gDOC& $0.858$/\underline{$\mathbf{0.530}$} & $0.860$/\underline{$\mathbf{0.585}$}  &$0.848$/\underline{$0.617$}  & $0.615$/$0.510$\\
     GOOD (own) & $0.868$/$0.505$  & $0.860$/$0.569$  & $0.848$/$\mathbf{0.622}$  & $0.227$/$\mathbf{0.647}$ \\

\bottomrule
\end{tabular}

    \caption{The test accuarcy/AUROC results for each OOD detector, model, and dataset combination in the semi-supervised classification setting. The best AUROC score for each GNN and dataset is marked in bold. The value of the OOD method used for GOOD is underlined.}
    \label{tab:semi-supervised}
\end{table}

\end{document}